\documentclass[lettersize,journal]{IEEEtran}
\usepackage{amsmath,amsfonts}
\usepackage{algorithmic}
\usepackage{algorithm}
\usepackage{array}
\usepackage[caption=false,font=normalsize,labelfont=sf,textfont=sf]{subfig}
\usepackage{textcomp}
\usepackage{stfloats}
\usepackage{url}
\usepackage{wrapfig}
\usepackage{listings}
\usepackage{multirow}
\usepackage{amssymb}% http://ctan.org/pkg/amssymb
\usepackage{pifont}% http://ctan.org/pkg/pifont
\usepackage{array}
\usepackage[export]{adjustbox}
\usepackage{xcolor}
\usepackage{comment}
\usepackage{arydshln}
\usepackage{makecell}
\usepackage{cancel}
\usepackage{verbatim}
\usepackage{graphicx}
\usepackage{cite}
\usepackage{hyperref}
\usepackage{amssymb}% http://ctan.org/pkg/amssymb
\usepackage{pifont}% http://ctan.org/pkg/pifont
\newcommand{\cmark}{\ding{51}}%
\newcommand{\xmark}{\ding{55}}%

\begin{document}

\title{Noise-Level Diffusion Guidance: Well Begun is Half Done}

\author{Harvey Mannering, Zhiwu Huang, and Adam Prugel-Bennett 
        % <-this % stops a space
\thanks{All authors belong to the Vision, Learning, and Control (VLC) Research Group, at the University of Southampton (email: hgm1g14@soton.ac.uk, Zhiwu.Huang@soton.ac.uk, apb@ecs.soton.ac.uk).}% <-this % stops a space
\thanks{This work has been submitted to the IEEE for possible publication.
Copyright may be transferred without notice, after which this version may
no longer be accessible. Manuscript received September 29, 2025.}}

%\markboth{IEEE TRANSACTIONS ON CIRCUITS AND SYSTEMS FOR VIDEO TECHNOLOGY, VOL. X, NO. Y, AUGUST 2025}%


\maketitle

\begin{abstract}
Diffusion models have achieved state-of-the-art image generation. {However, the random Gaussian noise used to start the diffusion process influences the final output, causing variations in image quality and prompt adherence.} % While diffusion-level guidance (DLG) methods like classifier-free guidance are well-explored, noise-level guidance (NLG) remains underutilized despite its potential to enhance image quality and prompt adherence. 
Existing noise-level optimization approaches generally rely on extra dataset construction, additional networks, or backpropagation-based optimization, limiting their practicality. In this paper, we propose Noise Level Guidance (NLG), a simple, efficient, and general noise-level optimization approach that refines initial noise by increasing the likelihood of its alignment with general guidance — requiring no additional training data, auxiliary networks, or backpropagation. The proposed NLG approach provides a unified framework generalizable to both conditional and unconditional diffusion models, accommodating various forms of diffusion-level guidance. Extensive experiments on five standard benchmarks demonstrate that our approach enhances output generation quality and input condition adherence. By seamlessly integrating with existing guidance methods while maintaining computational efficiency, our method establishes NLG as a practical and scalable enhancement to diffusion models. Code can be found at \href{https://github.com/harveymannering/NoiseLevelGuidance}{https://github.com/harveymannering/NoiseLevelGuidance}. %The code will be made publicly available.
\end{abstract}

\section{Introduction}
\label{sec:intro}

\IEEEPARstart{D}{iffusion} models have emerged as state-of-the-art generative models, demonstrating exceptional performance in synthesizing high-quality images across various applications. Their scalability to large datasets has made them particularly effective for tasks such as text-to-image generation. These models operate by progressively refining Gaussian noise over multiple denoising steps, ultimately producing coherent and visually compelling images. {However, as studied in many works \cite{mao2023guided,wang2024silent,qi2024not,li2024enhancing,ma2025scaling}, the quality of the final output is inherently influenced by the properties of the initial noise, with recent studies \cite{guo2024initno,eyring2024reno,ahn2024noise,sundaram2024cocono,wallace2023end,zhou2024golden} suggesting that noise-level optimization (NLO) can play a crucial role in enhancing image quality and improving adherence to conditioning signals. These techniques seek to optimize, refine, or select favorable noise distributions at an early-stage before the reverse diffusion process, enhancing overall image fidelity and coherence in generated outputs.}

An ideal NLO method should satisfy three key principles: simplicity, efficiency, and generality. First, it should be \emph{simple}, avoiding the need for additional dataset construction and auxiliary networks. Second, it should be \emph{efficient}, requiring no backpropagation and no complex verification through network inference. Third, it should be \emph{general}, applying across different types of diffusion models for unconditional and conditional image generation. However, to the best of our knowledge, as shown in Table~\ref{tab:NLO_comp}, no existing NLO approach fully meets all three essential criteria, highlighting a key gap in practical noise refinement techniques.

\begin{figure}[t]
   \centering

   \includegraphics[width=0.8\linewidth]{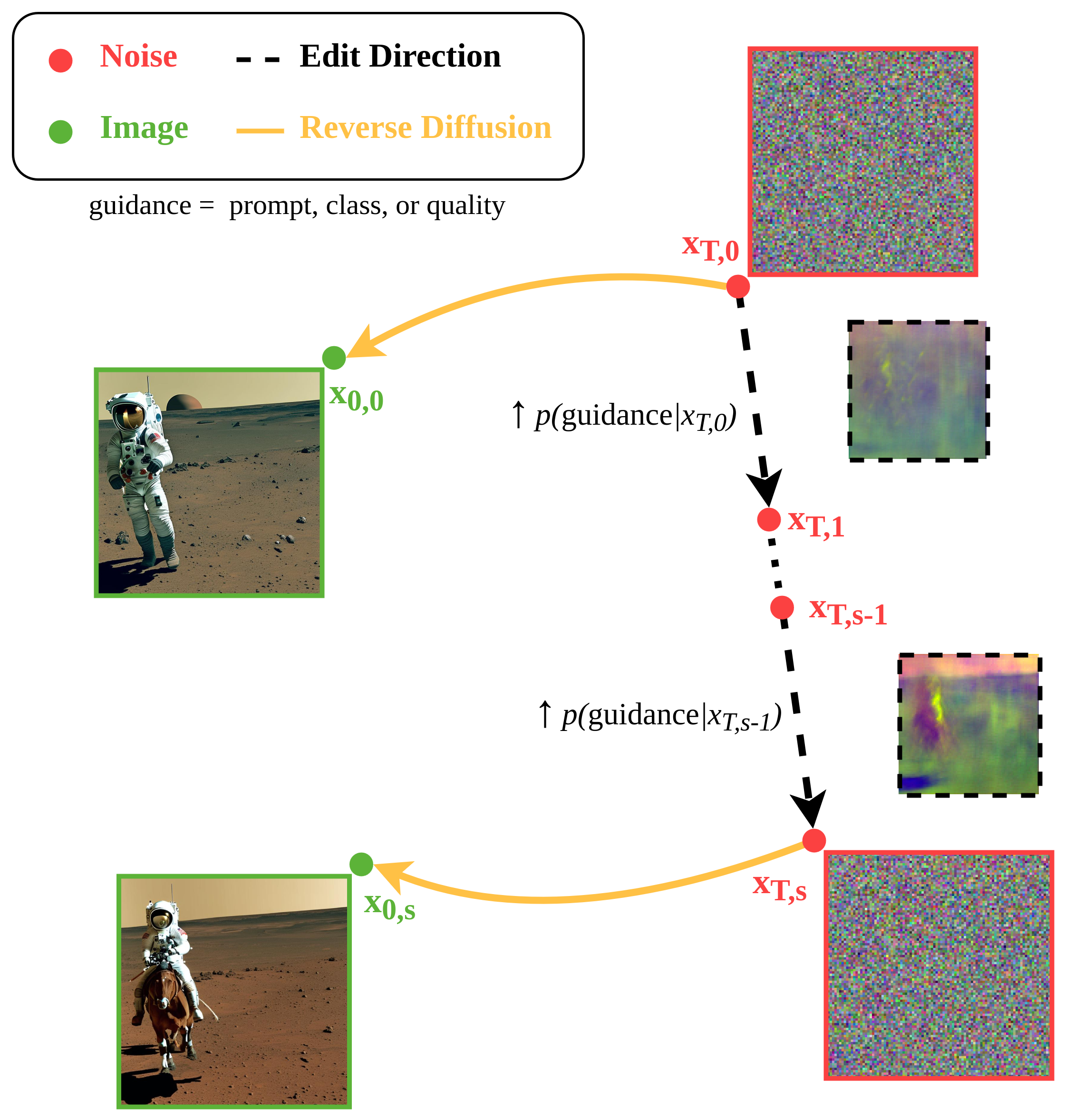}

\caption{Illustration of the proposed noise-level guidance (NLG) approach for aligning noise with \emph{general guidance} (e.g., a textual prompt, class label, or quality measure). To be efficient, NLG \emph{directly increases the likelihood of $p(\texttt{guidance}|x_T)$ with it's edit direction}, steering noise $x_T$ toward \emph{\texttt{guidance}}—e.g., ``\texttt{a photo of an astronaut riding a horse on Mars}’’—over $s$ forward steps. Edit directions are normalized for visualization. }

   \label{fig:method_core}
\end{figure}

To address this challenge, we propose a noise-level guidance (NLG) approach that is simple, efficient, and general. Instead of relying on computationally expensive search or backpropagation-based optimization procedures, the proposed NLG approach directly increases the likelihood of $p(\texttt{guidance}|\texttt{noise})$, moving \texttt{noise} in a direction that enhances alignment with \texttt{guidance}, which could be a textual prompt, class label, or quality measure. Fig.~\ref{fig:method_core} illustrates this process in the context of textual prompts. The key idea is to derive an edit direction from guidance-like mechanisms through a simple linear combination of diffusion model outputs under different conditions. This edit direction is applied iteratively to the initial noise, progressively steering it to better align with the provided target (either desired condition or quality measure) while preserving the stochastic nature of diffusion sampling.

The proposed NLG approach introduces a practical and computationally efficient alternative to existing NLO techniques, significantly expanding the capabilities of diffusion models with minimal overhead. Unlike prior approaches, it requires no dataset construction, no auxiliary networks, no backpropagation, and no verification. Furthermore, it generalizes well across unconditional and conditional image diffusion and rectified flow models, demonstrating consistent improvements in image quality and prompt adherence. We also explore downstream applications of NLG, including better alignment of generated images with text prompts and embedding desired properties into noise before generation.

In summary, our contributions are threefold:  

\begin{itemize}
    \item Simple and Efficient NLG: A lightweight noise refinement method requiring no additional datasets, extra networks, backpropagation, or verification.  
    \item Generalizable NLG: Generalizable across many unconditional and conditional image diffusion and rectified flow models without architectural changes.  
    % \item Complementary to DLG: \ZH{Reinforcing weak  diffusion-level guidance}, leading to improved diffusion models across standard benchmarks.  
    \item Comprehensive Evaluation: Experiments on five standard benchmarks show that the proposed NLG outperforms or matches SOTA methods, with \emph{4$\times$ faster speed} and \emph{3$\times$ lower memory usage compared to competing method InitNO \cite{guo2024initno}}, showing excellent \emph{generalization across both unconditional and conditional generation tasks with various types of diffusion-level guidance}.  
\end{itemize}

\begin{table}
\begin{center}
\caption{Summary of the proposed NLG and current noise-level optimization (NLO) methods. Simple: no additional datasets, no auxiliary networks. Efficient: no backpropagation, no verification by running the whole diffusion process. General: generalizable to both unconditional and conditional generation. $\dagger$: published paper, \xmark$^\#$: limited by single-step models due to gradient issue~\cite{eyring2024reno}, \xmark$^*$: highly inefficient (e.g., \cite{eyring2024reno} reported \cite{wallace2023end} requires 83.33 A100 days for inference on T2I-CompBench~\cite{huang2023t2i}) , \bcancel{\cmark}: efficient during inference, but requires an expensive training phase, \cmark$^*$: feasible in theory but not yet justified.}
 \label{tab:NLO_comp}
 \begin{tabular}{|p{3cm}|>{\centering\arraybackslash}p{1cm}|>{\centering\arraybackslash}p{1cm}|>{\centering\arraybackslash}p{1cm}|}  % Removed unnecessary vertical bar
   \hline 
   \textbf{Method} & Simple & Efficient & General \\
   \hline
   NoiseRefine~\cite{ahn2024noise} & \xmark & \multicolumn{1}{c|}{\bcancel{\cmark}} & \xmark \\
   Golden Noise \cite{zhou2024golden} & \xmark & \multicolumn{1}{c|}{\bcancel{\cmark}} & \xmark  \\
   Noise Hypernetworks \cite{eyring2025noise} & \xmark & \multicolumn{1}{c|}{\bcancel{\cmark}} & \cmark$^*$  \\
   \hline
   InitNO$^{\dagger}$ \cite{guo2024initno} & \cmark & \xmark & \xmark   \\
   CoCoNo~\cite{sundaram2024cocono} & \cmark & \xmark & \xmark \\
    \hline

   ReNO$^{\dagger}$ \cite{eyring2024reno} & \xmark & \xmark$^{\#}$ & \cmark$^*$ \\
 DOODL$^{\dagger}$~\cite{wallace2023end} & \xmark & \xmark$^*$ & \cmark$^*$ \\
 DNO$^{\dagger}$~\cite{tanginference} & \xmark & \xmark & \cmark$^*$ \\
    \hline

   Noise Selection~\cite{qi2024not} & \cmark & \xmark$^*$ & \cmark$^*$ \\
   Zero-Order Search$^{\dagger}$~\cite{ma2025scaling} & \xmark & \xmark  & \cmark$^*$   \\
    \hline

   \textbf{Proposed NLG} & \cmark & \cmark & \cmark \\
   \hline
 \end{tabular}
 \end{center}
\end{table}

\section{Related Work}
\label{sec:related_work}
% \ZH{It’s better to include a table in this section to highlight the key differences.}

\noindent\textbf{Diffusion.} Diffusion models \cite{sohl2015deep,ho2020denoising,song2020score} are trained to denoise images with varying levels of Gaussian noise. During inference, images can be generated by starting with pure Gaussian noise, and then gradually removing noise with the diffusion model until an image emerges.  Diffusion models currently offer state of the art performance in image generation \cite{karras2024analyzing,alemohammad2024self}.  Text-to-image models like Stable Diffusion \cite{Rombach_2022_CVPR}, Imagen \cite{saharia2022photorealistic}, and FLUX \cite{flux2024}, condition diffusion models on text embeddings obtained from models like CLIP \cite{radford2021learning}, enabling diverse image generation from text.  However, misalignment between generated images and prompts remains an issue.

To enhance outputs, many diffusion models use guidance.  Most commonly, classifier-free guidance \cite{ho2022classifier} uses a linear combination of a models unconditional and conditional output to increase the likelihood of a condition, and improve image quality. Follow-up work has since introduced a variety of refinements and alternative guidance mechanisms to enhance the control and quality of model outputs \cite{hong2023improving,brack2023sega,ahn2024self,karras2024guiding,kwon2025tcfg,chen2025normalized}. Diffusion guidance lies at the core of our method and will be discussed in more detail in Section~\ref{sec:method}.

% \paragraph{Initial Noise.}
\noindent\textbf{Noise-Level Optimization.}
% \ZH{Better reorganize the logic for this part. For example, we could travel along ``simple", ``efficient", and ``general" to discuss these existing NLO methods.} 
Recent work shows that the noise used to generate an image significantly impacts its semantics and quality \cite{mao2023guided,wang2024silent,qi2024not,li2024enhancing}.  Following this, several techniques have been suggested to find good initial noises for generation.  One approach generates a large set of images, analyzes them, and creates a noise database linked to specific properties \cite{mao2023semantic,wang2024silent}.  For example, \cite{mao2023semantic} create a layout-to-image synthesis model, collecting pixel blocks (random noise chunks tied to semantic concepts) and assembling them for controlled generation. These approaches are fast and efficient at inference time, but first require the construction and subsequent storage of a large database of noise.  

Methods such as Golden Noise \cite{zhou2024golden}, NoiseRefine \cite{ahn2024noise}, and Noise Hypernetworks \cite{eyring2025noise} train a neural networks to refine Gaussian noise which improves image quality. Training data is typically gathered using the diffusion model itself. The network is then trained to refine noise, leading to higher quality images. While efficient at inference time, they require dataset generation and model training. In contrast, our method needs no auxiliary networks, dataset generation, or training, making it simpler.

Sample-based selection methods \cite{qi2024not,ma2025scaling} involve generating images using a large number of candidate noises and then selecting the best one.  While general and scalable, these methods significantly increase inference time, making them inefficient.

Optimization-based techniques iteratively perturbs noise by backpropgating through the diffusion model \cite{wallace2023end,eyring2024reno,samuel2024generating,guo2024initno,sundaram2024cocono,tanginference}. InitNO \cite{guo2024initno} and CoCoNO \cite{sundaram2024cocono} define their loss function over attention maps to mitigate subject mixing in text-to-image models.  These methods require no additional components but demand high memory and are limited to text-to-image applications, making them neither efficient nor general. DOODL \cite{wallace2023end}, ReNO \cite{eyring2024reno}, and DNO \cite{tanginference} run rewards models to grade generated images (using models like CLIP \cite{radford2021learning} or ImageReward \cite{xu2024imagereward}) and then backpropagate through the whole system to adjust the noise. DNO supports non-differentiable rewards but still requires backpropagation through the diffusion model itself. The main drawback of these methods is that they must store the computational graph, which demands substantial memory and makes them impractical. For example, ReNO is limited to single-step diffusion models due to this restriction.  Our method perturbs noise using the model’s own output, eliminating the need for backpropagation. Furthermore, DOODL, ReNO, and DNO all require additional networks, while we do not.  This makes our approach simpler and more efficient.

\section{Proposed Approach}
\label{sec:method}

\noindent\textbf{Preliminaries.}
Diffusion models typically define a mapping between a normal distribution and a distribution approximated by a dataset over several steps. \cite{song2020score} found that diffusion models learn to approximate $\nabla_{x_t}\log p(x_t)$ at each timestep $t$, where $x_t$ is a noisy image.  At every step, a diffusion model's output will be scaled and subtracted from the $x_t$. %We primarily consider $\epsilon$-prediction due to its alignment with the score-function-based interpretation of diffusion, although our approach can also be applied to $v$-prediction and $x$-prediction with simple reparameterization.

Conditional generation is therefore given as $\nabla_{x_t}\log p(x_t|y)$ where $y$ is a condition (e.g. a text prompt).  Applying Bayes rule to this term yields:
\begin{equation}
  \nabla_{x_t}\log p(x_t|y) =  \nabla_{x_t}\log p(x_t) + \nabla_{x_t}\log p(y|x_t).
  \label{eq:bayes}
\end{equation}
Thus, conditional generation combines unconditional generation and a direction in pixel space that increases the likelihood of the condition $y$. Using this insight, diffusion-level guidance methods can be used to steer the diffusion process towards a certain goal. For example, classifier guidance \cite{dhariwal2021diffusion} controls the likelihood of $y$ by scaling the second term in equation~\eqref{eq:bayes} with a weight $w$.
%\begin{equation}
%  \nabla_{x_t}\log p(x_t) + w\nabla_{x_t}\log p(y|x_t)
  %\label{eq:classifier_guidance}
%\end{equation}
However, this requires an auxiliary classifier $p(y|x_t)$ which must be trained to be robust to noise.  Classifier-free guidance (CFG) \cite{ho2022classifier} removes the need for this external classifier by applying Bayes rule again to the second term, enabling users to increase the likelihood of a given condition using a combination of conditional and unconditional outputs from the diffusion model:
\begin{equation}
  w\nabla_{x_t}\log p(x_t|y) + (1-w)\nabla_{x_t}\log p(x_t).
  \label{eq:cfg}
\end{equation}
Autoguidance~\cite{karras2024guiding} (AutoG) uses a different guidance method to improve image quality:
\begin{equation}
  w D_1(x|y) + (1 - w)D_0(x|y),
  \label{eq:autoguide}
\end{equation}
where $D_1$ and $D_0$ are high-quality and low-quality diffusion models, respectively.  This allows guidance to be applied to unconditional generation (when $y=\emptyset$) and currently achieves state of the art image quality generation. 
%This family of methods, which we refer to as diffusion-level guidance, is applied at every timestep.  Our noise-level guidance on the other hand, is applied only at the starting timestep. One key advantage with only using the starting timestep is that the distribution is well known. At all other timesteps, our distribution will be a combination of a complexe image distirbution and a normal distirbuion. On the other, at the first time step, our distirbution is purely a normal distribution. This allows for use to take extra steps to prevent guidance points going our of distribution, which is a common problem with CFG.
% All guidance formulas also have the effect of improving image quality.

\noindent\textbf{Overview.} Inspired by these diffusion-level guidance formulas, we propose a noise-level guidance (NLG) approach that alters our starting noise such that the likelihood of a given condition is increased for conditional generation and the image quality is enhanced for unconditional generation.

For conditional generation, we use the difference between conditional and unconditional noise predictions as an edit direction, applying it to the initial noise to better align the final image with the input condition. As shown in Fig.~\ref{fig:method_core}, in the context of text-to-image models, applying an edit direction to our initial noise can improve alignment between the input prompt and the generated image.  In other words, the likelihood $p(y|n)$ can be increased, where $y$ is our text prompt and $n$ is our initial noise.  We also see that this works by adding subtle structure to the noise which corresponds to the given input prompt. For unconditional generation, the general nature of NLG allows us to leverage AutoG to find an edit direction that improves image quality, seamlessly integrating with the AutoG architecture.

Unlike diffusion-level guidance methods, we only apply a guidance direction at the first timestep where we know the distribution is a standard normal.  This allows us to take extra steps to prevent the random noise from going out of distribution. Specifically, we clip the edit direction’s length to constrain large edit directions, add slight Gaussian noise for stabilization, and normalize the latent to maintain the expected Gaussian noise magnitude at each iteration of the NLG approach.

\subsection{Aligning Noise with General Guidance}
\label{sec:method_aligning}

\noindent\textbf{Conditional Guidance}. 
% \ZH{@Harvey, this part should focus on prompt-, class-, and quality-based (autoguidance) conditional guidance? NLG+CFG? NLG+AutoG?} 
Rearranging equation~\eqref{eq:bayes} allows us to find a direction $d$ in latent space that increases the likelihood of a condition $y$ by taking the difference between conditional and unconditional generation:

\begin{align}
\refstepcounter{equation}
  d &= D_1(x_t|y) - D_0(x_t) \tag{\theequation a} \label{eq:implicit_classifier_a} \\
    &\approx \nabla_{x_t}\log p(y|x_t) \tag{\theequation b} \label{eq:implicit_classifier_b}
\end{align}

\noindent where $D_1$ and $D_0$ are diffusion models for conditional and unconditional generation.  If we use CFG, which is currently the most common form of guidance, then we set $D_1=D_0$.  However, when using AutoG, $D_1$ and $D_0$ are distinct models.  Diffusion starts by first sampling random noise $n\sim \mathcal{N}(0,I)$.  Our method then sets $x_t=n$ and subtracts $d$ from our random noise, perturbing $n$ in a way that increases the likelihood of $y$.  We then repeat this process for a total of $s$ steps.

To fully generalize our edit direction $d$, we can also include a second condition.
\begin{equation}
  d = D_1(x_t|y_1) - D_0(x_t|y_0),
\label{eq:edit_direction_general}
\end{equation}
where $y_0$ and $y_1$ are the negative and positive conditions, decreasing and increasing their respective likelihoods.

\noindent\textbf{Unconditional Guidance}. The formula for AutoG is given in equation~\eqref{eq:autoguide}, where $D_1$ and $D_0$ correspond to high quality and a lower quality diffusion models.  We can consider the scenario where we unify unconditional models $D_1$ and $D_0$ into a single model $D$, which is conditioned on either a high or low quality label (i.e. $y_h=\text{high quality}$  and $y_l=\text{low quality}$):
\begin{equation} D(x|y) = \begin{cases} D_1(x) & \text { if } y=y_h, \\ D_0(x) & \text { else }. \end{cases}
\label{eq:single_model}
\end{equation}
In this case, unconditional AutoG approximates:

\begin{align}
\refstepcounter{equation}
D_w(x) & = wD_1(x) + (1-w)D_0(x) \tag{\theequation a} \label{eq:autoguide_full_a} \\
& = wD(x|y_h) + (1-w)D(x|y_l) \tag{\theequation b} \label{eq:autoguide_full_b}\\
& \approx w\nabla \log p(x|y_h) + (1-w)\nabla \log p(x|y_l) \tag{\theequation c} \label{eq:autoguide_full_c}\\
& = w(\nabla \log p(y_h|x) +\nabla \log p(x)) \notag \\
& \quad\quad\ - (1-w)(\nabla \log p(y_l|x) + \nabla \log p(x)) \tag{\theequation d} \label{eq:autoguide_full_d} \\
& = w\nabla \log p(y_h|x) + (1-w)\nabla \log p(y_l|x) \notag \\
& \quad\quad\ + \nabla \log p(x). \tag{\theequation e} \label{eq:autoguide_full_e}
\end{align}

\noindent Provided $w>1$, we can see that the first term of equation~\eqref{eq:autoguide_full_e} increases the likelihood of high quality images, the second term decreases the likelihood of low quality images, and the third term is unconditional image generation. This provides some insight into why AutoG produces such high quality images, as we can see that AutoG is pushing generated images towards high quality generations, or at least high quality according to the implicit classifier $p(y|x)$.

Parallels can be drawn here between this implicit classifier and a discriminator.  If instead of predicting high or low quality, we were predicting real or fake images (a distinct but in practice is very similar tasks), then our implicit classifier would be a discriminator.  Unlike GANs \cite{goodfellow2014generative}, this use of a discriminator would not require adversarial training, and therefore should not suffer the same stability issues.  This ``Discriminator-Free Guidance'' set up is roughly equivalent to SIMS \cite{alemohammad2024self}.

If we want to align noise for unconditional generation, AutoG can therefore be used.  In this case our edit direction would be calculated as:

\begin{align}
\refstepcounter{equation}
  d &= D_1(x) - D_0(x) \tag{\theequation a} \label{eq:autoguide_3a} \\
    &\approx \nabla \log p(y_h|x) - \nabla \log p(y_l|x). \tag{\theequation b} \label{eq:autoguide_3b}
\end{align}

We can see that in this instance, $d$ will be a direction in latent space that increases the likelihood of a high quality image and decreases the likelihood of a low quality image.

Repeatedly applying the edit direction $d$, calculated either with equation~\eqref{eq:implicit_classifier_a}, \eqref{eq:edit_direction_general}, or \eqref{eq:autoguide_full_a}, is the core of our noise aligning method, however, this risks pushing noise $n$ out-of-distribution of the standard normal, resulting in low quality generated images.  The next section details three methods that help ensure that our noise can better stay in-distribution.

\begin{figure}[t]
   \centering
    \begin{minipage}[b]{0.48\textwidth}
    
    \centering
    \texttt{{PROMPT:} "A man stands on the platform with his back turned to the train"}
  
  \vspace{1mm}
  {\setlength{\tabcolsep}{1.5pt}% Reduce space between columns
  \rotatebox{90}{\makebox[0mm][c]{~ w/ Clip ~~~ wo/ Clip ~~}}
  \begin{tabular}{@{}c c c c c@{}}
    % Column headers (first empty cell for the row-end text)
    \textbf{s=0} & \textbf{s=1} &  \textbf{s=5} & \textbf{s=10} & \\[1ex]
    % First row of images with green text at the end
    \includegraphics[width=0.215\textwidth]{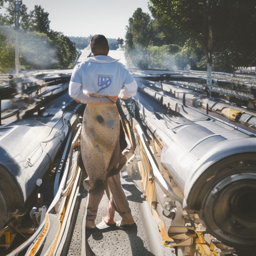} &
    \includegraphics[width=0.215\textwidth]{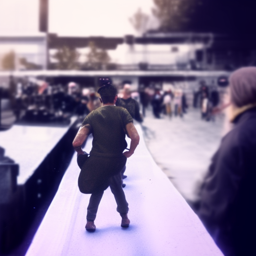} &
    \includegraphics[width=0.215\textwidth]{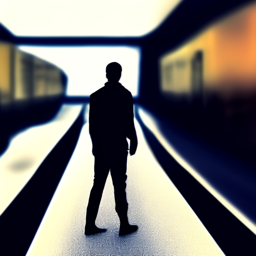} &
    \includegraphics[width=0.215\textwidth]{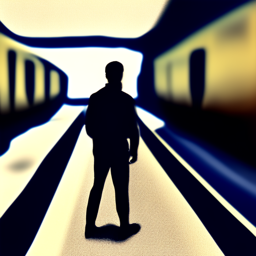} \\
    \includegraphics[width=0.215\textwidth]{Figures/nlg_images/clipping/573261_r0.png} &
    \includegraphics[width=0.215\textwidth]{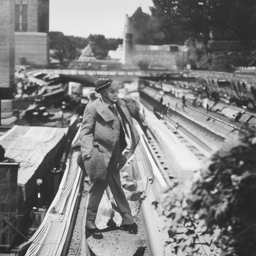} &
    \includegraphics[width=0.215\textwidth]{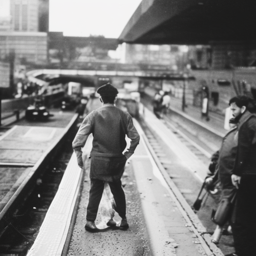} &
    \includegraphics[width=0.215\textwidth]{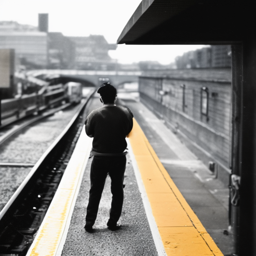} \\
  \end{tabular}
  }
  \end{minipage}
  
  \begin{minipage}[b]{0.48\textwidth}
    %--- Second Grid ---
  \centering
  \vspace{3.5ex}  
  \texttt{PROMPT: "A bus stopped near the sidewalk at a bus"}
   
  \vspace{1mm}
  {\setlength{\tabcolsep}{1.5pt}%
  \rotatebox{90}{\makebox[0mm][c]{~w/ Clip ~~~ wo/ Clip ~~}}
  \begin{tabular}{@{}c c c c c@{}}
    \textbf{s=0} & \textbf{s=1} & \textbf{s=5} & \textbf{s=10} & \\ 
    [1ex]
    % Second row of images with red text at the end
    \includegraphics[width=0.215\textwidth]{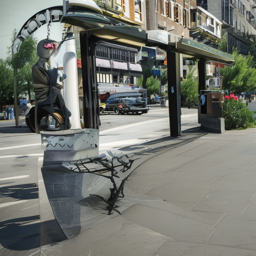} &
    \includegraphics[width=0.215\textwidth]{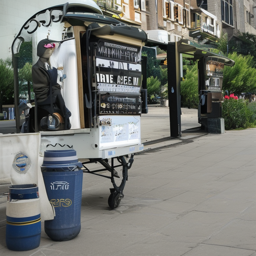} &
    \includegraphics[width=0.215\textwidth]{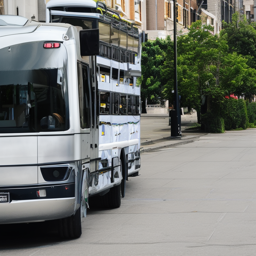} &
    \includegraphics[width=0.215\textwidth]{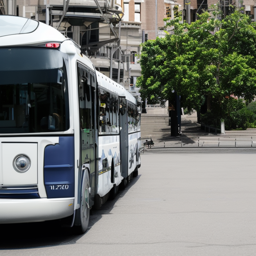} \\
    \includegraphics[width=0.215\textwidth]{Figures/nlg_images/clipping/234608_r0.png} &
    \includegraphics[width=0.215\textwidth]{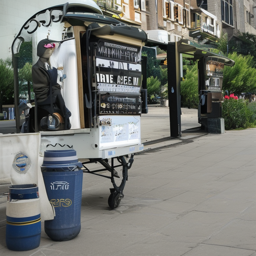} &
    \includegraphics[width=0.215\textwidth]{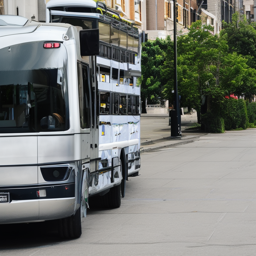} &
    \includegraphics[width=0.215\textwidth]{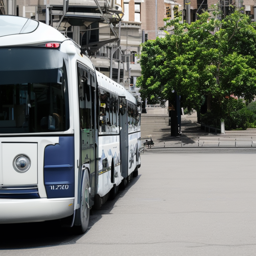} \\
  \end{tabular}
  }
  \end{minipage}
    %--- First Grid ---
  
  \vspace{3.5ex} % Vertical space between grids

   \caption{Images generated with and without direction clipping.  Stable Diffusion v2.1 was used for generation.  We see that direction clipping improves quality for images with large edit directions (the ``man'' images) and leaves images unchanged when edit directions are small (the ``bus'' images).}
   \label{fig:clipping}
\end{figure}

\subsection{Out-of-Distributions Errors}
\label{sec:method_errors}

Repeatedly subtracting a direction $d$ from random noise can quickly push the noise out of distribution.  To avoid this, we clip the length of $d$, add a small amount of noise, and normalize the latent to be the expected magnitude of Gaussian noise at each iteration.

\begin{figure}[t]
   \begin{center}
   \includegraphics[width=0.6\linewidth]{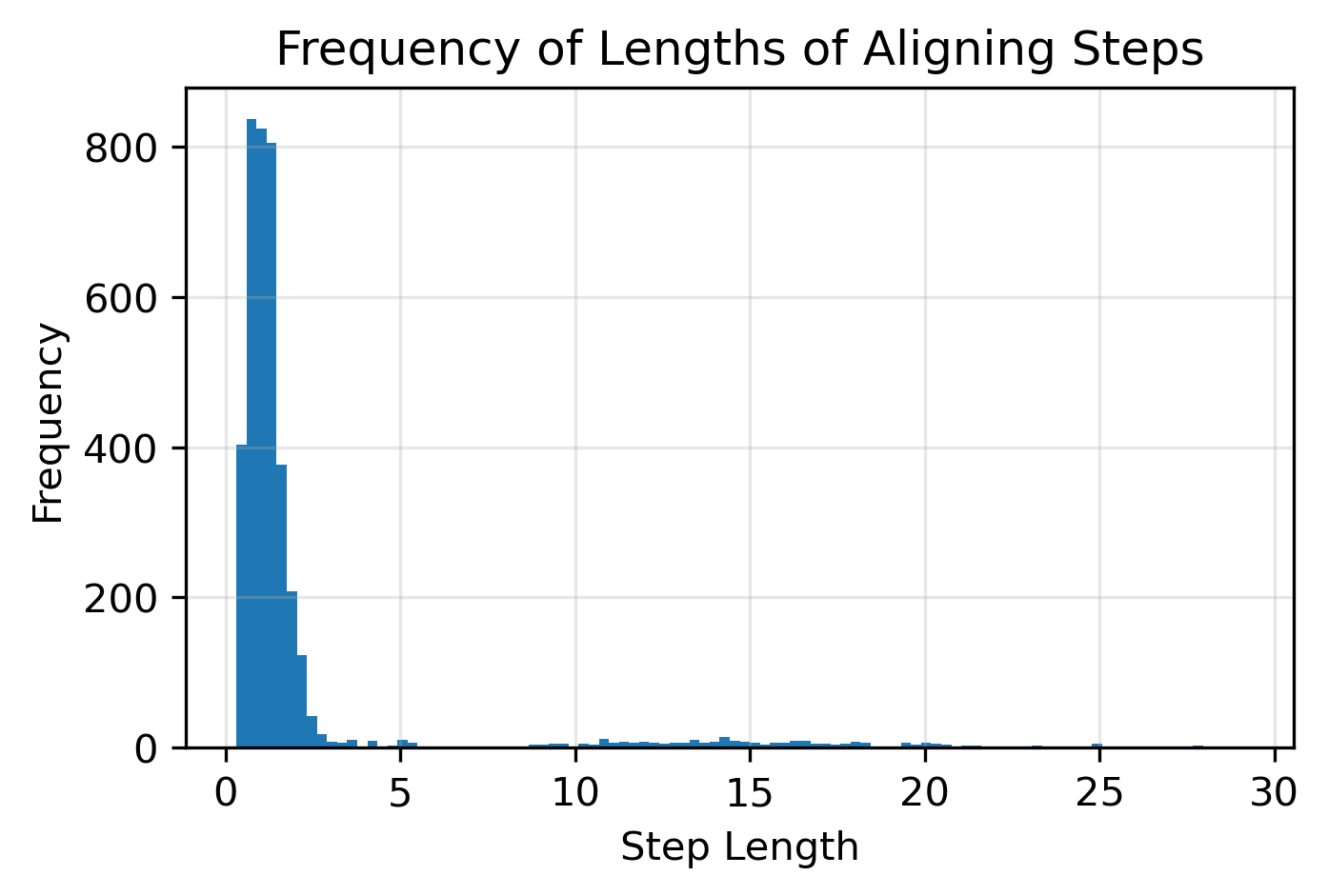}
   \end{center}
   \caption{{100 noises were refined on Stable Diffusion v2.1 using prompts from MS COCO 2014 validation set.  We do this for 20 aligning steps and record the length of every step.  The histogram shows the lengths of the recorded steps. Most steps are clustered in the 0 to 5 range, however, a small number of steps have much larger lengths.}}
   \label{fig:histogram}
\end{figure}

\noindent\textbf{Direction Clipping.}  Fig.~\ref{fig:clipping} shows two images as their latents are progressively aligned for more and more steps.  The ``bus" image produces optimal result at around 10 iterations, but the ``man" image after only 1 iteration. This is caused by drastically different lengths of $d$, which results in the ``man" image going out of distribution quicker than the ``bus" image. To negate this, we follow \cite{eyring2024reno} and clip the direction, meaning that if $\|d\|$ is above a certain threshold $\tau$, we normalize it such that $\|d\|=\tau$.  As seen in Fig.~\ref{fig:clipping}, this negates the negative effects of large edit directions (seen in a small number of images), while making little or no different to images with smaller edit directions.  Fig.~\ref{fig:histogram} shows the lengths of edit directions for 100 images over the course of 20 aligning steps.  We see that most steps have lengths of between 0 and 5, however, a small number of refining steps will have more extreme lengths.  These more extreme lengths cause our noise to go out of distribution more quickly, leading to artifacts in generated images.  Since only a small number of images are effected by extreme edit directions, metrics may not reflect the benefit of direction clipping.

\noindent\textbf{Additional Noise.} To further prevent out-of-distribution errors, we follow \cite{wallace2023end} and introduce a small amount of Gaussian noise at every iteration. Empirically, we find that $\mathcal{N}(0,10^{-3} \cdot I)$ achieves a good balance between image quality and the likelihood of condition $y$.

\noindent\textbf{Normalization.} Normally distributed points in a high-dimensional space sit roughly on a hypersphere with an approximate radius of $\sqrt{a}$, where $a$ is the number of dimensions.  For a full proof of this see \cite{zheng2024noisediffusion}.  To avoid our noise $n$ having the incorrect expected magnitude, we can therefore multiply it by $\frac{\sqrt{a}}{\|n\|}$ at every iteration of algorithm.
In cases where the starting distribution is not the standard normal, but instead $\mathcal{N}(0,\sigma_{max}^2I)$, our normalizing constant becomes $\frac{\sigma_{max}\sqrt{a}}{\|n\|}$. However, in most cases $\sigma_{max}=1$.

\noindent\textbf{Full Algorithm.} For the complete noise aligning process see Algorithm~\ref{alg:alg_main}.  In this algorithm we calculate the direction using two conditions $y_1$ and $y_0$.  If we set $y_0=\emptyset$, then we obtain equation~\eqref{eq:implicit_classifier_a}. If $y_1=y_0=\emptyset$, then we recover equation~\eqref{eq:autoguide_full_a}. 

\begin{algorithm}[H]
\caption{Proposed NLG Approach}\label{alg:alg_main}
\begin{algorithmic}
\REQUIRE $s$ (Total Aligning Steps), $D$ (Diffusion Models), $y$ (Conditions), $\tau$ (Clipping Threshold), $a$ (Dimensions of Latent Space), $l$ (Additional Noise Level), $\sigma_{max}$ (Starting Noise Level)
\STATE $n \gets \mathcal{N}(0,\sigma_{max}^2I)$
\FOR{$i=0~$ \textbf{to} $~s~$}
    \STATE $d \gets D_1(n|y_1) - D_0(n|y_0)$ %\ZH{Can this step be more general?}
    \STATE $d \gets \text{NormClip}(d, \tau)$
    \STATE $n \gets n - d + \mathcal{N}(0,lI)$
    \STATE $n \gets \frac{\sigma_{max}\sqrt{a}}{\|n\|}n$
\ENDFOR
\end{algorithmic}
\end{algorithm}

\section{Experiments}
\label{sec:experiments}

\subsection{Analysis on Proposed Approach}
\label{sec:experiments_analysis}

\begin{figure*}
\begin{center}
\begin{tabular}{ccc}
\includegraphics[height=4.0cm]{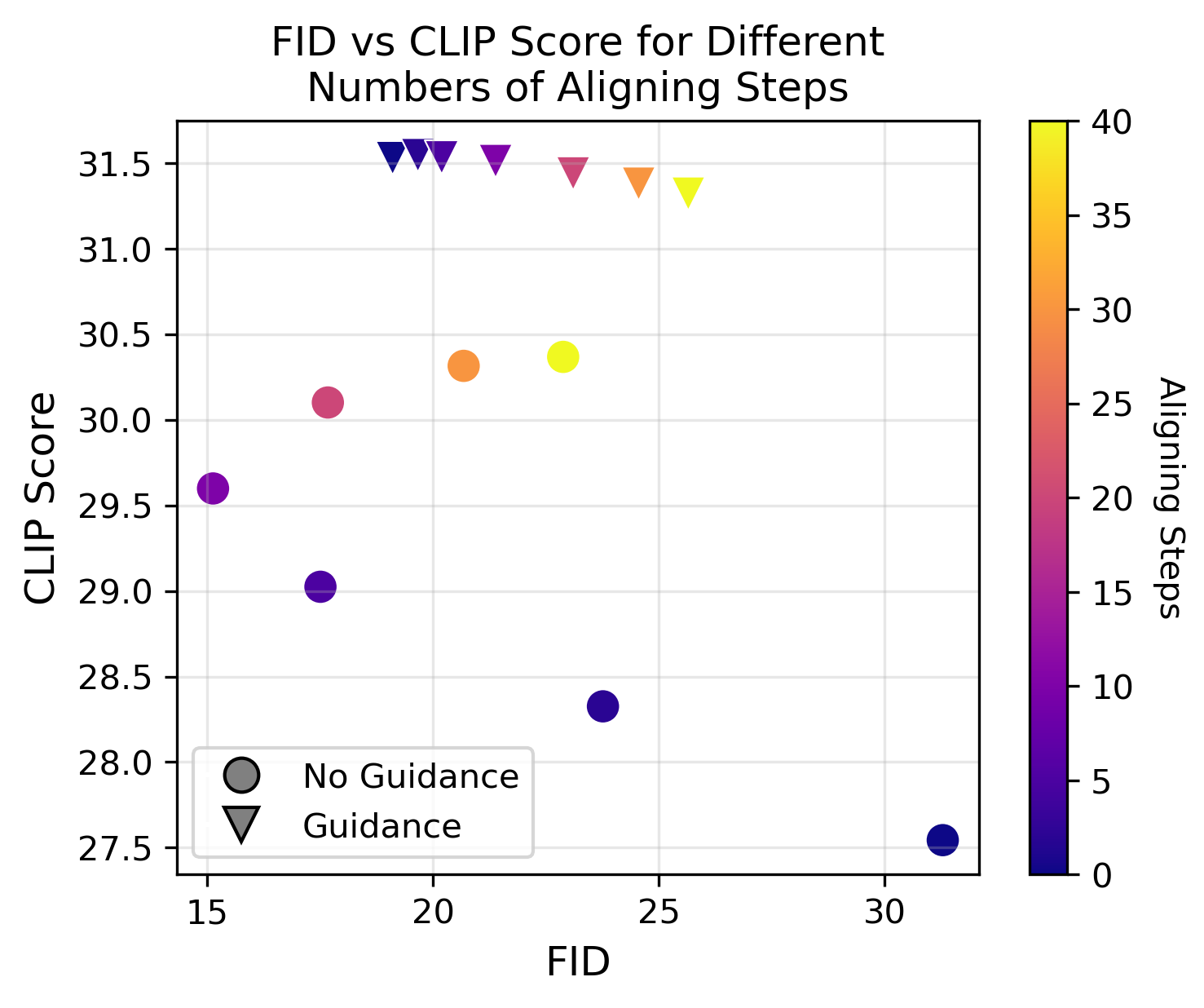} &
\includegraphics[height=4.0cm]{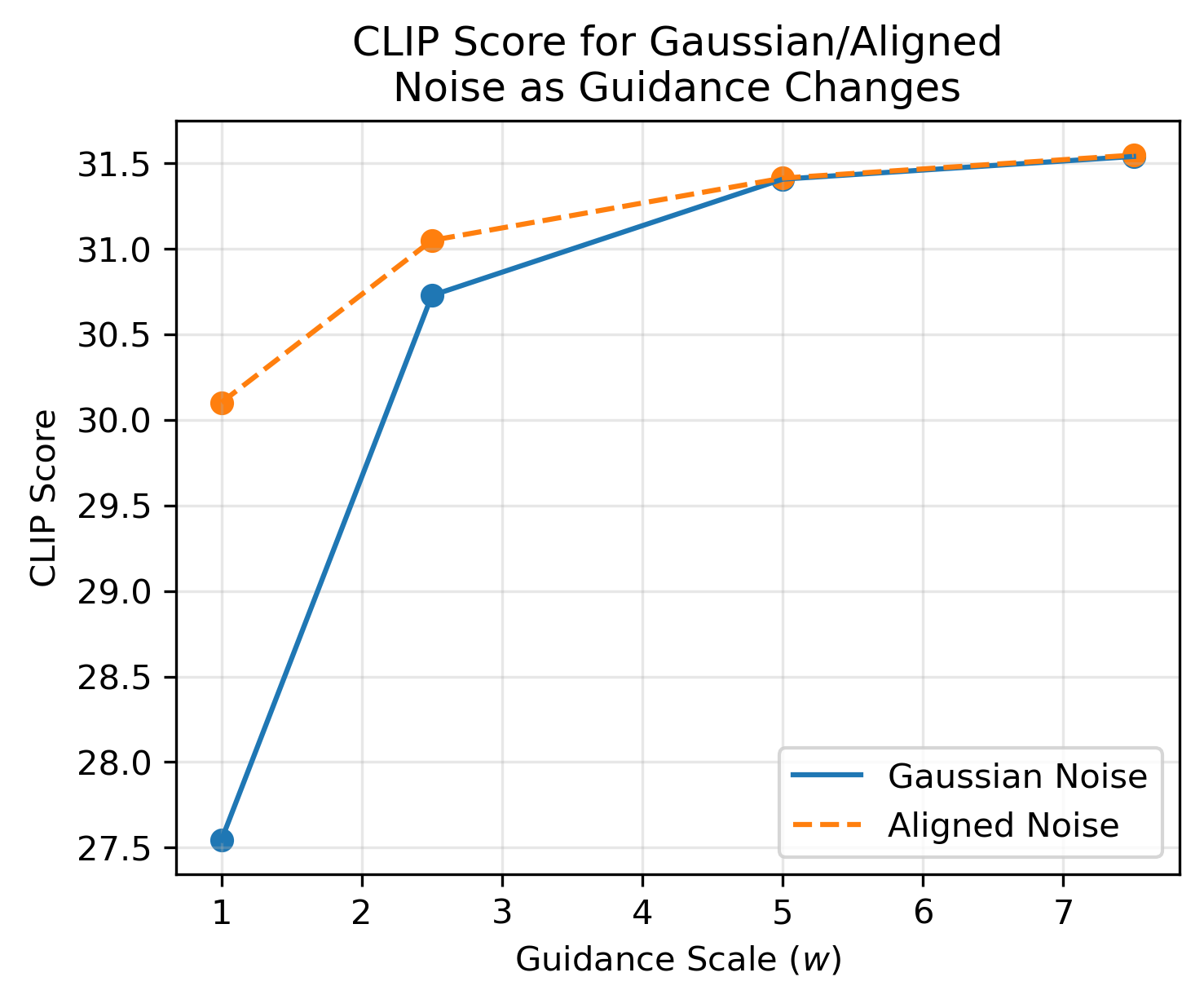}&
\includegraphics[height=4.0cm]{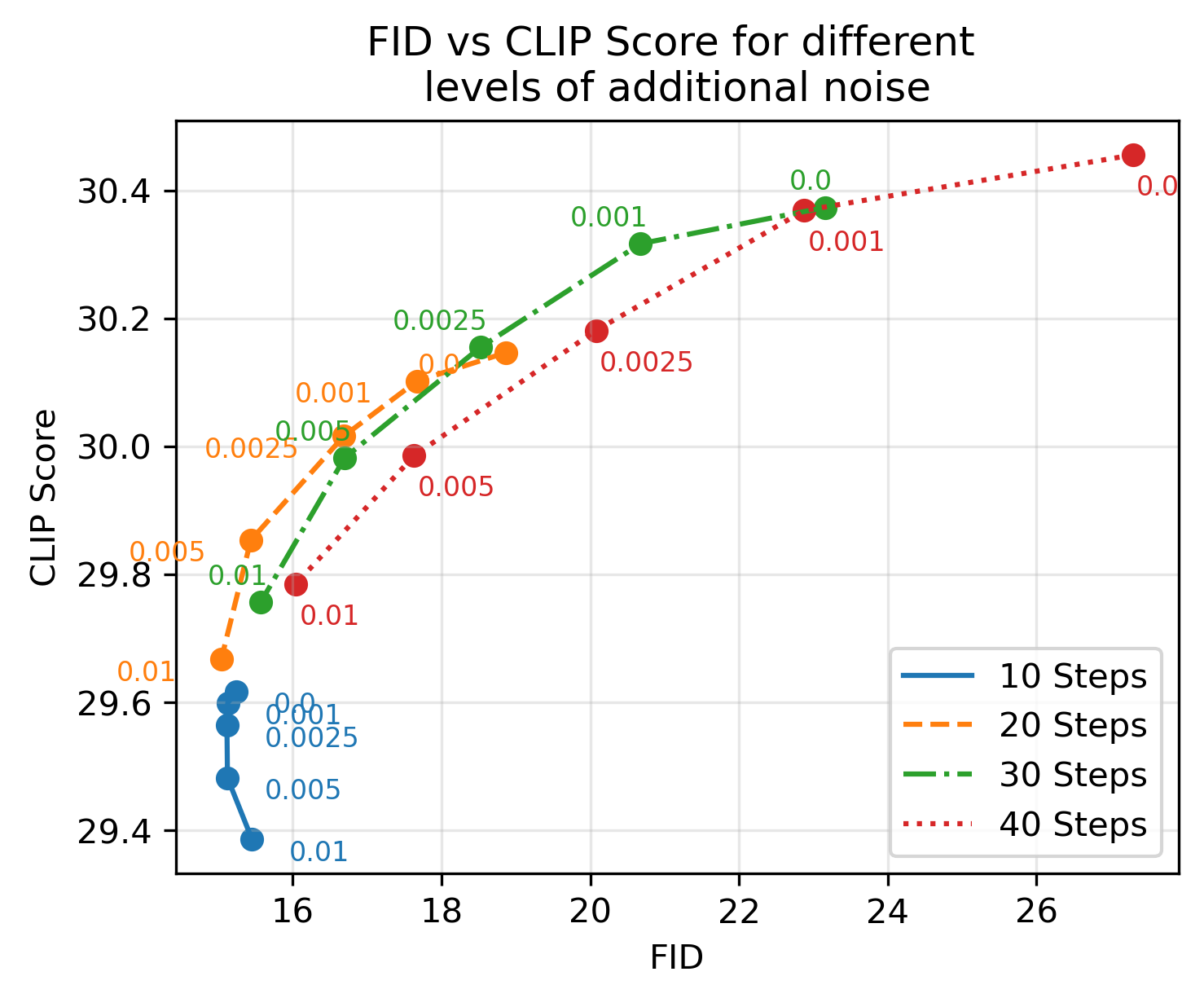} \\
(a) Aligning Steps $s$ & (b) CFG Scales $w$ & (c) Noise Level $l$
\end{tabular}
\end{center}
\caption{\textbf{(a)} Impact of using different numbers of NLG steps at CFG guidance scales $w=7.5$ and $w=1.0$, \textbf{(b)} CLIP scores of NLG for different CFG scales ($w=1, 2.5, 5, 7.5$),  \textbf{(c)} Trade-off between CLIP score and FID for different numbers of NLG steps and noise levels for NLG. No guidance is used ($w=1.0$). {Noise levels $l$ are shown next to each point with small digits.}} %\ZH{We should highlight Fig3c, which seems a big picture of showing our overall results.}
\label{fig:graphs}
\end{figure*}

We first look at the performance of our approach across a variety of settings, analyzing the impact of hyperparameters and measuring overall performance.  Unless stated otherwise we use Stable Diffusion v2.1 \cite{rombach2022high} with 20 inference steps.  Following \cite{ahn2024noise}, we source prompts from the MS COCO 2014 validation set \cite{lin2014microsoft} and generate 10k images per measurement. We assess image quality using FID \cite{heusel2017gans} and Inception Score (IS) \cite{salimans2016improved}, while prompt-image alignment is measured with CLIP Score \cite{radford2021learning}. Since the aim of our direction $d$ is to increase the likelihood of a given text prompt, we should expect the main impact of our method to be an increase in CLIP Score. The clipping threshold is $\tau=0.5$.

\noindent\textbf{Aligning Steps.} We perform a grid search over aligning steps $s = {0, 2, 5, 10, 20, 30, 40}$, testing two generation settings: without CFG ($w=1$) and with CFG ($w=7.5$, Stable Diffusion's default). Results are shown in Fig.~\ref{fig:graphs}a.  In the first setting, NLG improves both FID and CLIP Score up to $s=10$, after which increasing $s$ improves CLIP Score but degrades FID. In the second setting, NLG provides minimal improvement, with $s=2$ showing a small increase in CLIP Score at the cost of FID. This suggests NLG is most beneficial when no guidance is used during generation.

\noindent\textbf{Guidance Scale.} Guidance scale therefore seems to be a key determinant of when our method will be useful.  In no-guidance scenarios, NLG improves image-text alignment, but in high CFG guidance settings, the benefits are minimal (this mirrors previous NLO work \cite{ahn2024noise}). We test CFG scales $w = 1, 2.5, 5, 7.5$ using the optimal $s$ values for $w=1$ and $w=7.5$ ($s=20$ and $s=2$, respectively).  For guidance scales in between $1$ and $7.5$, we interpolate the number of aligning steps ($16$ steps for $w=2.5$ and $9$ steps for $w=5$). Results in Fig.~\ref{fig:graphs}b show that NLG improves alignment for both no guidance and low guidance ($w=2.5$) settings, but not for high guidance. However, for images with low CLIP Scores, NLG can still be beneficial even with high guidance (see Section~\ref{sec:nlg_applications}).

\noindent\textbf{Additional Noise Level.}  We test the impact of adding Gaussian noise at each aligning step, with noise levels $l = {0, 0.001, 0.0025, 0.005, 0.01}$. The noise levels $l$ corresponds to the variance of the normal distribution from which the addition noise is sampled.  We perform this experiment in the no diffusion-level guidance setting, and repeat it for different numbers of aligning steps. In Fig.~\ref{fig:graphs}c, we again observe a clear trade-off between FID and CLIP Score, however, additional noise appears effective at balancing these two goals (a similar effect was observed in \cite{zheng2024noisediffusion}).  For example, at 20 steps adding addition noise with $l=0.001$ decreases FID from 18.86 to 17.67 (6.3\% improvement) while only decreasing CLIP Score from 30.15 to 30.10 (0.2\% deterioration).  We choose $s=20$ and $l=0.001$ for subsequent experiments as it best balances FID, CLIP Score, and number of function evaluations.

\noindent\textbf{Normalization.} In Table \ref{tab:ablation} we ablate normalizing the latent after each aligning step.  The results show that normalization is necessary for high-quality generations. We attribute the sharp increase all metrics to the normalizing step's ability to prevent the noise going too far out of distribution.

\renewcommand{\arraystretch}{1}

\begin{table}
  \begin{center}
  \caption{Ablation of the normalization step in our NLG approach. Result obtained using 10k samples on Stable Diffusion v2.1.}
  \label{tab:ablation}
  \begin{tabular}{|c|c|c|c|}
    \hline 
    Normalization&FID $\downarrow$&IS $\uparrow$& CLIP Score $\uparrow$\\
    \hline
    \xmark & 156.06 & 6.32  & 27.72 \\
    \cmark & \textbf{17.67}  & \textbf{32.75} & \textbf{30.10} \\
    \hline
  \end{tabular}
  \end{center}
\end{table}

\begin{figure*}[htbp]
   \centering
  
  %--- Second Grid ---
  \vspace{1mm}
  {\setlength{\tabcolsep}{0.5pt}%
  
  \rotatebox{90}{\makebox[0mm][c]{ ~ Ours ~~~~~~~~~ $\mathcal{N}(0,I)$ ~~~~~~~~~~~~~~~~~}}
  \begin{tabular}{@{}c c c c c c c@{}}
    \begin{minipage}[b]{1.75cm}
    \scriptsize\centering
    \texttt{"A black Honda motorcycle parked in front of a garage."}
    \end{minipage} & 
    \begin{minipage}[b]{1.75cm}
    \scriptsize\centering
    \texttt{"A room with blue walls and a white sink and door."}
    \end{minipage} &
    \begin{minipage}[b]{1.75cm}
    \scriptsize\centering
    \texttt{"A car that seems to be parked illegally behind a legally parked car"}
    \end{minipage} & 
    \begin{minipage}[b]{1.75cm}
    \scriptsize\centering
    \texttt{"A large passenger airplane flying through the air."}
    \end{minipage} & 
    \begin{minipage}[b]{1.75cm}
    \scriptsize\centering
    \texttt{"The home office space seems to be very cluttered."}
    \end{minipage} & 
    \begin{minipage}[b]{1.75cm}
    \scriptsize\centering
    \texttt{"A cute kitten is sitting in a dish on a table."}
    \end{minipage} & 
    \begin{minipage}[b]{1.75cm}
    \scriptsize\centering
    \texttt{"An open food container box with four unknown food items."}
    \end{minipage} \\ [1ex] 
    
    % Second row of images with red text at the end
    
    \includegraphics[width=0.12\textwidth]{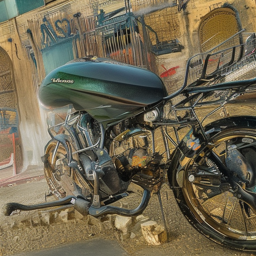} &
    \includegraphics[width=0.12\textwidth]{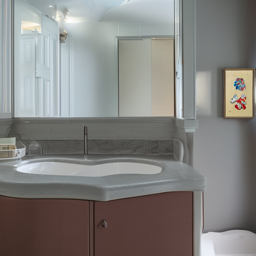} &
    \includegraphics[width=0.12\textwidth]{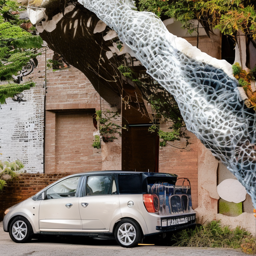} &
    \includegraphics[width=0.12\textwidth]{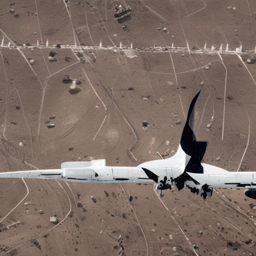}
    & \includegraphics[width=0.12\textwidth]{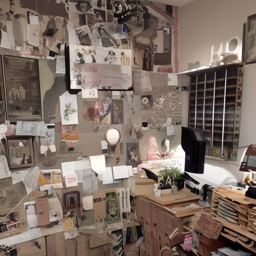}&
    \includegraphics[width=0.12\textwidth]{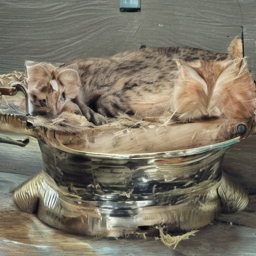}&
    \includegraphics[width=0.12\textwidth]{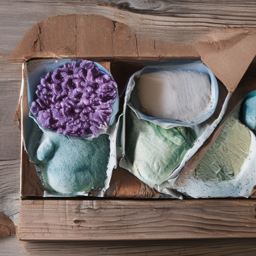}\\
    
    \includegraphics[width=0.12\textwidth]{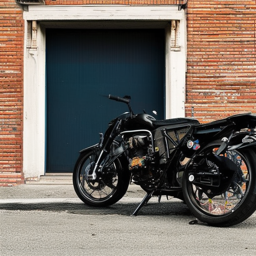} &
    \includegraphics[width=0.12\textwidth]{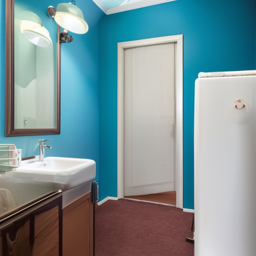} &
    \includegraphics[width=0.12\textwidth]{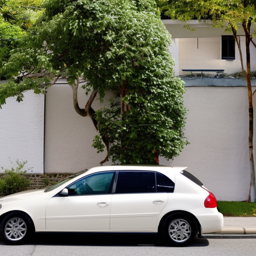} &
    \includegraphics[width=0.12\textwidth]{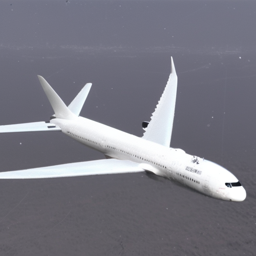}
    &
    \includegraphics[width=0.12\textwidth]{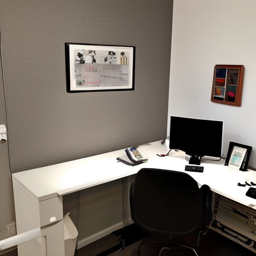}
    &    \includegraphics[width=0.12\textwidth]{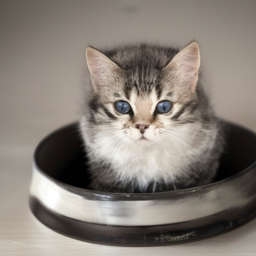}
    &
    \includegraphics[width=0.12\textwidth]{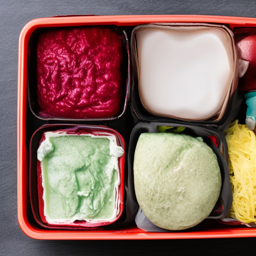}\\
  \end{tabular}
  }
  
   \caption{Qualitative results for Stable Diffusion v2.1 both with and without our NLG.  During generation, no guidance is used ($w=1$).  Prompts used for generation are sourced from MS COCO and are shown at the top of each column.}
   \label{fig:t2i_qualitative}
\end{figure*}

\begin{figure}[t]
   \begin{center}
  
  %--- Second Grid ---
  % \rotatebox{90}{\makebox[0mm][c]{\textbf{\underline{Stable Diffusion v1.5}~~~~~~~~~~~~~~~~}}}
  % \rotatebox{90}{\makebox[0mm][c]{ ~~ Ours ~~~~~~~ $\mathcal{N}(0,I)$ ~~~~~~~~~~~~~~~~}}
\vspace{1mm}
  {\setlength{\tabcolsep}{0.5pt}%
  
  \rotatebox{90}{\makebox[0mm][c]{ ~~~ Ours ~~~~~~~~ $\mathcal{N}(0,I)$ ~~~~~~~~~~~~~~~~}}
  \begin{tabular}{@{}c c c@{}}
    % Second row of images with red text at the end
    \begin{minipage}[b]{1.5cm}
    \scriptsize\centering
    \texttt{"A parking meter between two cars on a city street."}
    \end{minipage} & 
    \begin{minipage}[b]{1.5cm}
    \scriptsize\centering
    \texttt{"A large white dead polar bear on display in a museum."}
    \end{minipage} &
    \begin{minipage}[b]{1.5cm}
    \scriptsize\centering
    \texttt{"a brown teddy bear is sitting on a green bed"}
    \end{minipage}\\ [1ex] 
    \includegraphics[width=0.12\textwidth]{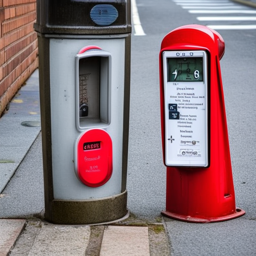}&
    \includegraphics[width=0.12\textwidth]{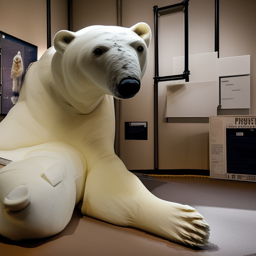}&
    \includegraphics[width=0.12\textwidth]{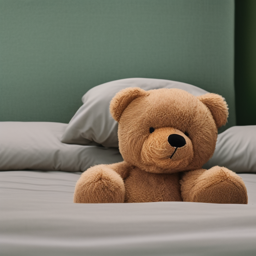}\\
    \includegraphics[width=0.12\textwidth]{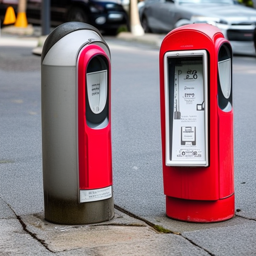}
    &    \includegraphics[width=0.12\textwidth]{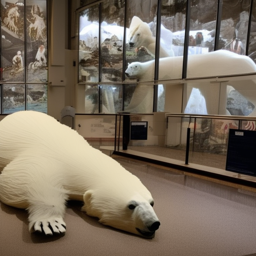}
    &
    \includegraphics[width=0.12\textwidth]{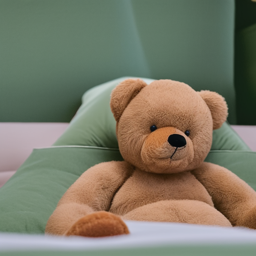}\\
  \end{tabular}
  }
  \end{center}
   \caption{Qualitative results for Stable Diffusion v2.1 using CFG, both with (i.e., Ours) and without our NLG (i.e., $\mathcal{N}(0,I)$). We use $w=7.5$ for all images.  Prompts used for generation are sourced from MS COCO and are shown at the top of each column.}
   \label{fig:t2i_qualitative_cfg}
   \vspace{-0.3cm}
\end{figure}

\noindent\textbf{Qualitative Results.} We present qualitative results in Fig.~\ref{fig:t2i_qualitative} (without CFG) and Fig.~\ref{fig:t2i_qualitative_cfg} (with CFG) for Stable Diffusion v2.1. These results demonstrate that aligning noise with our NLG can improve both image-text alignment and image quality.

\renewcommand{\arraystretch}{0.25}

\renewcommand{\arraystretch}{1}

\noindent\textbf{User Study.}
We design an anonymized user study to determine if our method is improving image quality and text alignment, and if so, by how much.  We conducted the survey using Qualtrics, an online survey platform, and, following the InitNO \cite{guo2024initno} evaluation protocol, collected responses from 11 participants (for a total of 2,200 responses).  Participants were made up of University staff and post-graduate researchers.  Each question presents the user with two images generated using Stable Diffusion v2.1 and the accompanying prompt used to generate the images. One image will correspond to the baseline model, and the other image corresponds to exactly same generation setting, except that the noise has been aligned using NLG prior to generation. The order of the images displayed to the user is randomized. For each pair of images the user is asked ``Which image looks more realistic?'' and ``Which image better aligns with the text prompt?''.  For each question they can then select either, image A, image B, or choose that both images are equal.

We collect responses for two settings, generated with CFG ($w=7.5$) and without CFG ($w=1$). All hyperparameter settings are inherited from our core experiments on Stable Diffusion v2.1 (i.e. 20 inference steps, 2 and 20 aligning steps used when we do and do not use CFG, respectively).  For each setting we compare 50 pairs of images.  We then counted the winning images for both questions.  Win rates for each setting are shown in Table \ref{tab:user_study}. We applied a one-tailed sign test ($\alpha=0.05$) to assess whether our NLG method is preferred over the Gaussian noise baseline. Without CFG, our method shows a statistically significant preference over Gaussian noise for both questions. With CFG enabled, our aligned noise maintains a higher win rate than Gaussian noise in both cases. 

\begin{table*}[t]
  \begin{center}
  \caption{Human evaluation results comparing Gaussian noise and our aligned noise on perceived realism (left) and text-prompt alignment (right), without and with CFG ($w=7.5$). Stable Diffusion v2.1 is used as our base model.}
  \label{tab:user_study}
  \begin{tabular}{|c|c|c|c|c|c|c|}
    \hline 
    &
    \multicolumn{3}{c|}{\thead{\footnotesize\textbf{Which image looks} \\ \footnotesize\textbf{more realistic?}}} 
    & \multicolumn{3}{c|}{\thead{\footnotesize\textbf{Which image better aligns} \\ \footnotesize\textbf{with the prompt?}}} \\
    \hline
    CFG & Gaussian Noise & NLG (Ours) & Equal 
             & Gaussian Noise & NLG (Ours) & Equal \\
    \hline
    \xmark & 6\% & 86\% & 8\% & 8\% & 64\% & 28\% \\
    \cmark & 16\% & 26\% & 58\% & 0\% & 6\% & 94\% \\
    \hline
  \end{tabular}
  \end{center}
\end{table*}

\noindent\textbf{{T2I-CompBench.}} T2I-CompBench \cite{huang2023t2i} is a collection of 6000 prompts that are challenging for text-to-image models.  These include prompts which describe different object attributes (colour, shape, and texture), prompts which describe spatial and non-spatial relations between objects, and complex prompts. Additional neural networks are then used to assess how well images align with the prompts used to generate them.

We therefore benchmark NLG on T2I-CompBench so we can better understand how our method handles more challenging prompts.  We use Stable Diffusion v2.1 with 20 inference steps.  We benchmark both with and without CFG during inference.  When CFG is used, we set $w=7.5$.  When NLG is used, we set $l=0.001$ and $\tau=0.5$, then set $s=2$ for the guidance setting and $s=20$ for the no guidance setting.

Results are shown in Table~\ref{tab:t2i-combench} where we see that, without CFG, our method increases metrics by an average of 0.04.  When CFG is used, metrics on average remain the same.  This is consistent with our prior findings that any gains in text-image alignment are mostly only observed when diffusion-level guidance is not used. 

\begin{table*}
  \begin{center}
  \caption{Benchmarking results on T2I-CompBench   \cite{huang2023t2i}. Stable diffusion v2.1 was used as the base model with 20 inference steps.  When guidance is used, we set $w=7.5$. We use 20 refining steps when no guidance (CFG) is used and 2 refining steps when guidance (CFG) is used.}
  \label{tab:t2i-combench}
  \begin{tabular}{|c|c|c|c|c|c|c|c|}
    \hline 
    CFG & Noise & Color $\uparrow$ & Shape  $\uparrow$ & Texture $\uparrow$ & Non-Spatial $\uparrow$ & Spatial $\uparrow$ & Complex $\uparrow$ \\
    \hline
    \multirow{ 2}{*}{\xmark} & Gaussian & 0.31 & 0.35 & 0.39 & 0.27 & 0.04 & 0.27  \\
     & NLG (Ours) & 0.41 & 0.34 & 0.37 & 0.29 & 0.10 & 0.37 \\
     \hline
     \multirow{ 2}{*}{\cmark} & Gaussian & 0.54 & 0.45 & 0.52 & 0.31 & 0.18  & 0.43 \\
     & NLG (Ours) & 0.53 & 0.44 & 0.53 & 0.31 & 0.18 & 0.44 \\
    \hline
  \end{tabular}
  \end{center}
\end{table*}

\subsection{State-of-the-art NLO Methods}
\label{sec:experiments_competing}
Next, we compare against two state-of-the-art methods for refining starting noise: InitNO \cite{guo2024initno} and NoiseRefine \cite{ahn2024noise}.  See supplementary material for further comparative work.

\begin{figure}[t]
   \centering
   \includegraphics[width=0.6\linewidth]{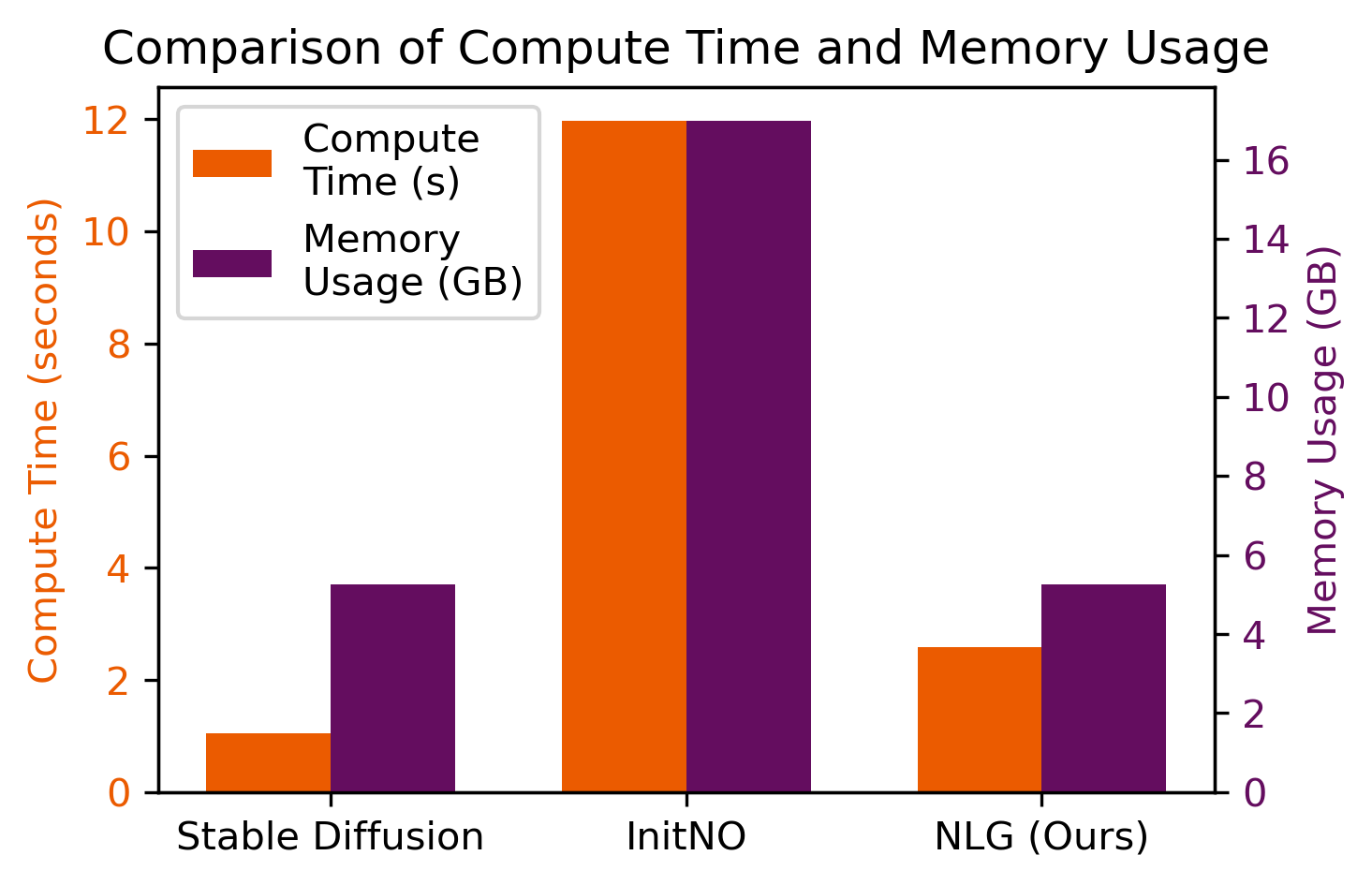}

   \caption{Comparison of average generation time per image and peak VRAM usage for Stable Diffusion v1.5, InitNO, and our NLG method on an NVIDIA A100 GPU. \emph{Same memory as Stable Diffusion, roughly $\times 4$ faster speed, $\times 3$ less memory than InitNO}}
   \label{fig:compute_bar_chart}
\end{figure}

\noindent\textbf{InitNO.} InitNO is a method for refining noise in text-to-image models to better align with the text prompts.  This is done by defining a loss function over the attention maps of the network, and then backpropagating to adjust the noise.  In contrast, our method does not require backpropagation.  As shown in Fig.~\ref{fig:compute_bar_chart}, this results in a \emph{$\times 4$ speedup and $\times 3$ reduction in peak GPU memory usage}.  We also do not rely on attention maps, allowing \emph{our method to be applied to unconditional generation} (see Section~\ref{sec:experiments_non_t2i}).

We benchmark against InitNO using the Attend-and-Excite dataset \cite{chefer2023attend}. The baseline model is Stable Diffusion v1.5 with 20 inference steps. We use $w=1.0$, $\tau=0.5$ and $l=0.001$. We generate 64 images per prompt and evaluate alignment with CLIP Text-Image and CLIP Text-Text similarity.  Since no reference images are available, we evaluate image quality using IS. We also include scores from human preference models ImageReward \cite{xu2024imagereward}, HPSv2.1 \cite{wu2023human}, and PickScore \cite{kirstain2023pick}.  Results in Table \ref{tab:initno} show that for this setting our NLG beats InitNO and the baseline Stable Diffusion model on nearly all standard evaluation metrics.

\begin{table*}
    \begin{center}
    \caption{Comparison with InitNO on the Attend-and-Excite dataset.  Prompts from all subsets are combined, leading to a total of 17664 images. Stable Diffusion v1.5 is used as the base model. See supplementals for more fine-grained breakdown of these metrics.}
  \label{tab:initno}
    \footnotesize
  \begin{tabular}{|c|c|>{\centering\arraybackslash}p{1.8cm}|>{\centering\arraybackslash}p{1.8cm}|c|c|c|}
    \hline 
    Method &IS $\uparrow$ & \thead{\footnotesize Text-Text \\Similarity $\uparrow$} & \thead{\footnotesize Image-Text \\ \footnotesize Similarity $\uparrow$} & ImageReward $\uparrow$ & HPSv2 $\uparrow$ & PickScore $\uparrow$ \\ %& \thead{\normalsize Generation \\ \normalsize Time $\downarrow$} & \thead{\normalsize Peak VRAM \\ \normalsize Usage $\downarrow$} \\
    \hline
    SD1.5 (Gaussian Noise) & \underline{16.31}  & 0.63 & 0.27 & -1.619 & 0.1889 & 19.54 \\ %& \textbf{1.0s} & \textbf{5.3 GB}  \\
    InitNO & 13.59 & \underline{0.68} & \underline{0.30} & \textbf{-0.571} & \underline{0.2098} & \underline{19.91} \\ % & 12.0s & 17.0 GB  \\
     NLG (Ours) & \textbf{19.53} & \textbf{0.73} & \textbf{0.32} & \underline{-0.727} & \textbf{0.2312} & \textbf{20.71} \\ %& \underline{2.6s} & \underline{5.3 GB} \\
    \hline
  \end{tabular}
  \end{center}
\end{table*}

\noindent\textbf{NoiseRefine.} We compare our NLG against NoiseRefine using IS, FID, CLIP Score, ImageReward, HPSv2.1, and PickScore on MS COCO. Since no code is available at time of writing, we use their reported numbers. Following their evaluation setup, we generate images using Stable Diffusion v2.1 with 20 inference steps.  Metrics are calculated using 30k images and no guidance is used when generating images.

\begin{table*}
  \begin{center}
  \caption{Comparison with NoiseRefine on the MS COCO 2014 validation dataset. Calculated using 30k samples. No CFG is used for these three methods. *\color{gray}{Reported results in \cite{ahn2024noise}}.} % \ZH{The reported digits of ours is for our balanced setting? What about NoiseRefine's digits? Do they from the same NoiseRefine setting? }}
  \label{tab:noise_refine}
  \begin{tabular}{|c|c|c|c|c|c|c|}
    \hline 
   Method &FID $\downarrow$&IS $\uparrow$& CLIP Score $\uparrow$ & ImageReward $\uparrow$ & HPSv2 $\uparrow$ & PickScore $\uparrow$\\
    \hline
    SD2.1 (Gaussian Noise) & 26.68 & 22.27  & 27.53 & -0.915 & 0.1973 & 19.83 \\
    NoiseRefine* & \color{gray}{11.39} & \color{gray}{35.73} & \color{gray}{30.27} & \color{gray}{0.338} & \color{gray}{0.2474} & \color{gray}{21.01} \\
    NLG (Ours) & {13.28} & {33.91} & {30.08} & {-0.183} & {0.2372} & {20.99} \\
    \hline
  \end{tabular}
  \end{center}
\end{table*}

Recall that NoiseRefine requires the generation of a large dataset (50k samples) which is then used to train an auxiliary model. At inference time, this model is then used to refine noise before being used for generation.  As shown in Table \ref{tab:noise_refine}, our NLG gets competitive performance while also requiring no dataset generation, no training, and no additional networks.

\begin{figure*}[http!]
  \begin{center}
  \begin{minipage}[b]{0.49\linewidth}
   \begin{center}
  \vspace{1mm}
  {\setlength{\tabcolsep}{1.5pt}%
  \rotatebox{90}{\makebox[0mm][c]{ \scriptsize ~~~ Ours ~~~~~~~~ $\mathcal{N}(0,I)$ }}
  \begin{tabular}{@{}c c c c c@{}}
    % First row of images with green text at the end
    \includegraphics[width=0.175\textwidth]{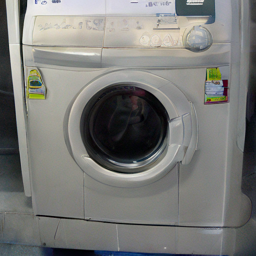} &
    \includegraphics[width=0.175\textwidth]{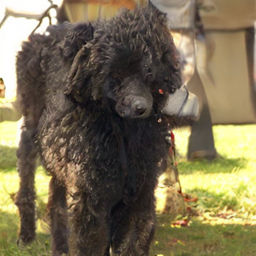}&
    \includegraphics[width=0.175\textwidth]{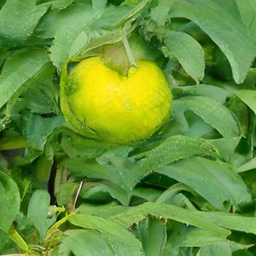}&
    \includegraphics[width=0.175\textwidth]{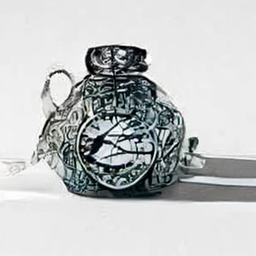} &
    \includegraphics[width=0.175\textwidth]{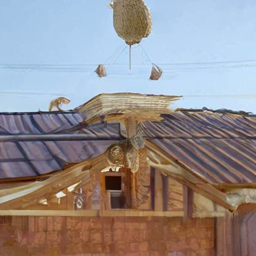}  \\
   \includegraphics[width=0.175\textwidth]{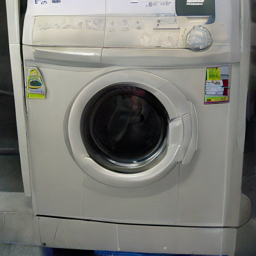} &
    \includegraphics[width=0.175\textwidth]{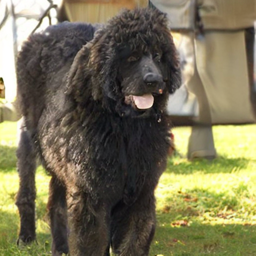}&
    \includegraphics[width=0.175\textwidth]{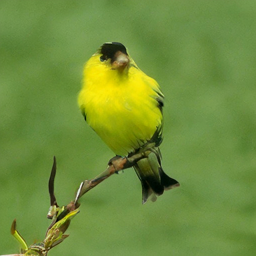}&
    \includegraphics[width=0.175\textwidth]{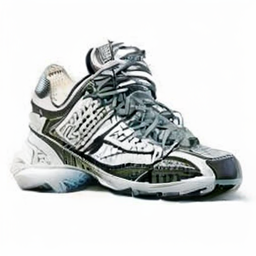} &
    \includegraphics[width=0.175\textwidth]{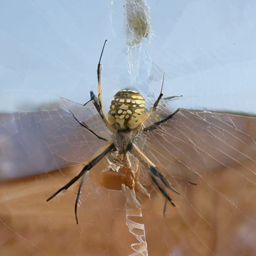}   \\
  \end{tabular}
  }
  \end{center}
 \begin{center}
      {\textbf{(a)} Unconditional Results ($w=1.00$)}
     \end{center}
    \end{minipage}
    \hfill
    \begin{minipage}[b]{0.49\linewidth}
   \begin{center}
  \vspace{1mm}
  {\setlength{\tabcolsep}{1.5pt}%
  \rotatebox{90}{\makebox[0mm][c]{\scriptsize ~~ Ours ~~~~~~~~ $\mathcal{N}(0,I)$ ~~~ }}
  \begin{tabular}{@{}c c c c c@{}}
    % Column headers (first empty cell for the row-end text)
    \tiny 1:Goldfish & \tiny 96:Toucan & \tiny 248:Husky & \tiny 883:Vase  & \tiny 178:Weimaraner \\[1ex]
    % First row of images
    \includegraphics[width=0.175\textwidth]{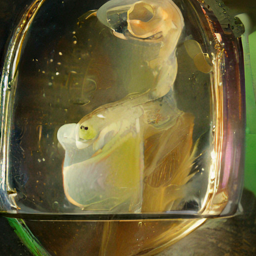} &
    \includegraphics[width=0.175\textwidth]{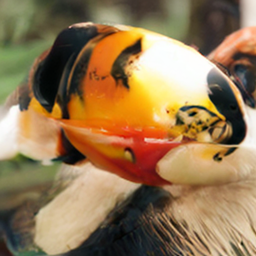}&
    \includegraphics[width=0.175\textwidth]{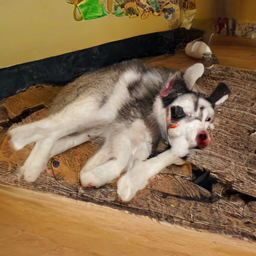}&
    \includegraphics[width=0.175\textwidth]{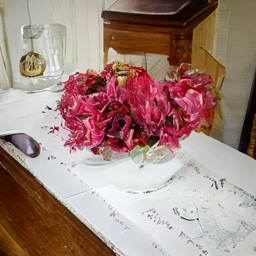} &
    \includegraphics[width=0.175\textwidth]{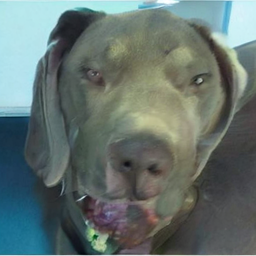}   \\
   \includegraphics[width=0.175\textwidth]{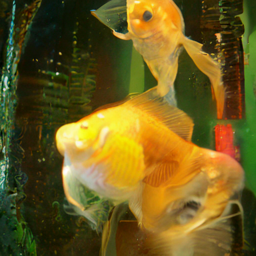} &
    \includegraphics[width=0.175\textwidth]{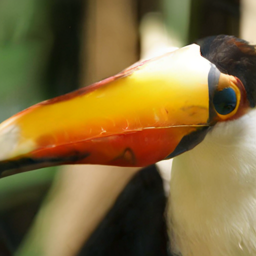}&
    \includegraphics[width=0.175\textwidth]{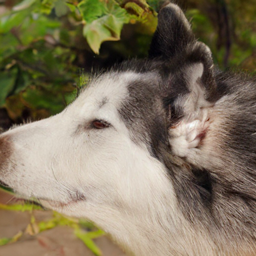}&
    \includegraphics[width=0.175\textwidth]{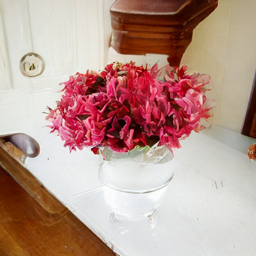}  &
    \includegraphics[width=0.175\textwidth]{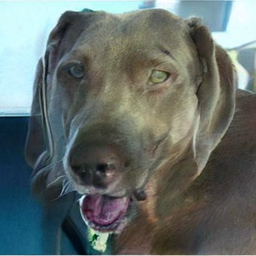}  \\
  \end{tabular}
  }
   \end{center}
    \begin{center}
    {\textbf{(b)} Conditional Results ($w=1.00$)}
     \end{center}
    \end{minipage}
    \\
    % Optional: add vertical space if needed
    \vspace{0.1cm}

    \begin{minipage}[b]{0.49\linewidth}
   \begin{center}
  \vspace{1mm}
  {\setlength{\tabcolsep}{1.5pt}%
  \rotatebox{90}{\makebox[0mm][c]{ \scriptsize ~~~ Ours ~~~~~~~~ $\mathcal{N}(0,I)$ }}
  \begin{tabular}{@{}c c c c c@{}}
    % First row of images with green text at the end
    \includegraphics[width=0.175\textwidth]{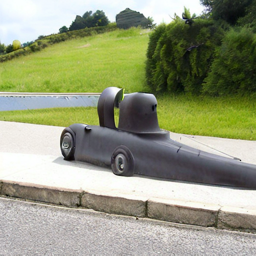} &
    \includegraphics[width=0.175\textwidth]{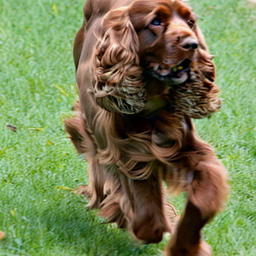}&
    \includegraphics[width=0.175\textwidth]{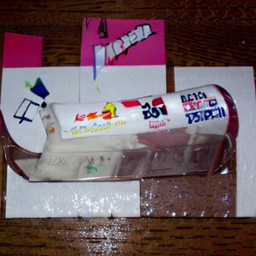}&
    \includegraphics[width=0.175\textwidth]{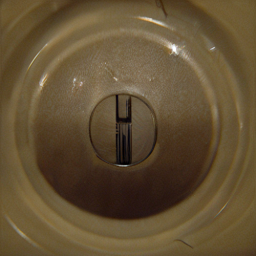} &
    \includegraphics[width=0.175\textwidth]{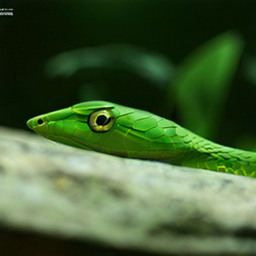}  \\
   \includegraphics[width=0.175\textwidth]{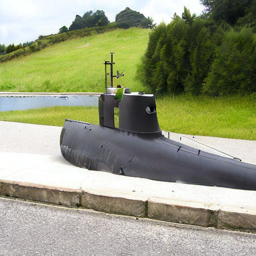} &
    \includegraphics[width=0.175\textwidth]{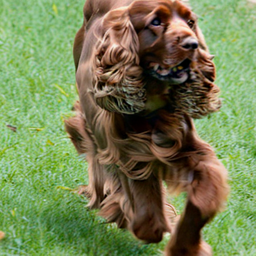}&
    \includegraphics[width=0.175\textwidth]{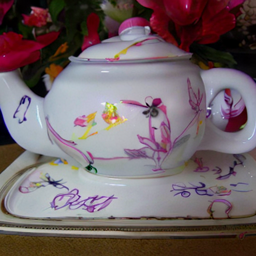}&
    \includegraphics[width=0.175\textwidth]{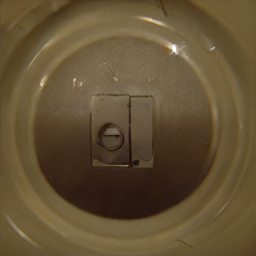} &
    \includegraphics[width=0.175\textwidth]{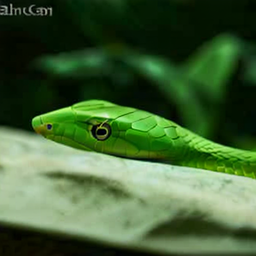}   \\
  \end{tabular}
  }
   \end{center} \begin{center}
      {\textbf{(c)} Unconditional Results ($w=2.90$)}
       \end{center}
    \end{minipage}
    \hfill
    \begin{minipage}[b]{0.49\linewidth}
   \begin{center}
  \vspace{1mm}
  {\setlength{\tabcolsep}{1.5pt}%
  \rotatebox{90}{\makebox[0mm][c]{  \scriptsize ~~ Ours ~~~~~~~~ $\mathcal{N}(0,I)$ ~~~}}
  \begin{tabular}{@{}c c c c c@{}}
    % Column headers (first empty cell for the row-end text)
    \tiny 863:Totem Pole & \tiny 231:Collie & \tiny 235:Alsatian & \tiny 679:Necklace  & \tiny 881:Upright Piano \\[1ex]
    % First row of images
    \includegraphics[width=0.175\textwidth]{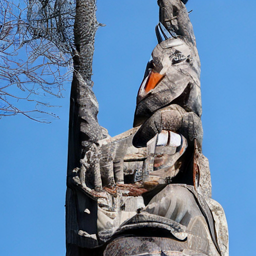} &
    \includegraphics[width=0.175\textwidth]{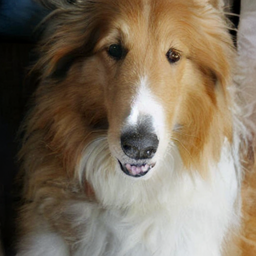}&
    \includegraphics[width=0.175\textwidth]{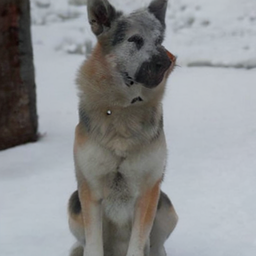}&
    \includegraphics[width=0.175\textwidth]{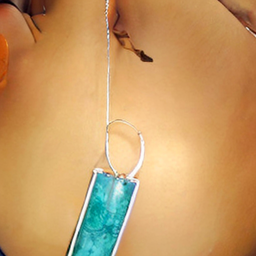} &
    \includegraphics[width=0.175\textwidth]{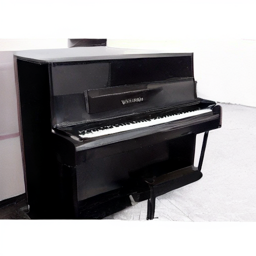}   \\
   \includegraphics[width=0.175\textwidth]{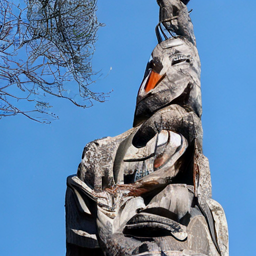} &
    \includegraphics[width=0.175\textwidth]{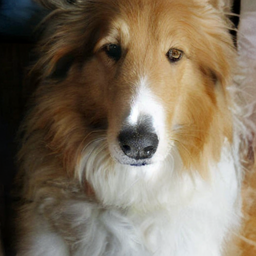}&
    \includegraphics[width=0.175\textwidth]{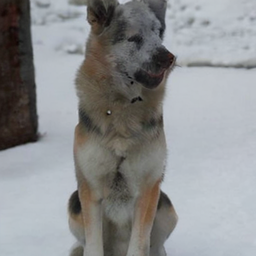}&
    \includegraphics[width=0.175\textwidth]{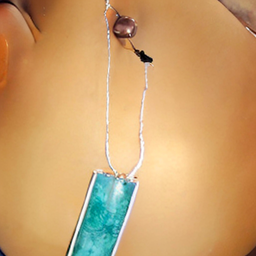}  &
    \includegraphics[width=0.175\textwidth]{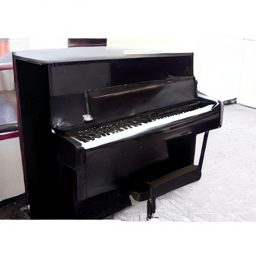}  \\
  \end{tabular}
  }
   \end{center}
    \begin{center}
    {\textbf{(d)} Conditional Results ($w=1.40$)}
  \end{center}   
    \end{minipage}
    \end{center}
    \caption{Qualitative results for EDM2 experiments.  For each setting we show images with and without aligned noise \textbf{(a)} Images generated in the unconditional setting with no guidance ($w=1.00$).  \textbf{(b)} Images generated in the conditional setting with no guidance ($w=1.00$).  Classes used for alignment and generation are shown above the columns. \textbf{(c)} Unconditional generation result using autoguidance ($w=1.40$  following EDM2 defaults). \textbf{(d)} Conditional generation result using autoguidance ($w=2.90$ following EDM2 defaults).}
    \vspace{-1.0cm}
    \label{fig:edm2_results}
\end{figure*}

\renewcommand{\arraystretch}{0.25}

\begin{figure*}[htbp]
   \centering
  
  %--- Second Grid ---
  \vspace{1mm}
  {\setlength{\tabcolsep}{0.5pt}%
  
  \rotatebox{90}{\makebox[0mm][c]{~~\textbf{\underline{Stable Diffusion v3.5}~~~~~~~~~~~~~~~~~~~~}}}
  \rotatebox{90}{\makebox[0mm][c]{ ~~ Ours ~~~~~~ $\mathcal{N}(0,I)$ ~~~~~~~~~~~~~~~~~}}
  \begin{tabular}{@{}c c c c c c c@{}}
    \begin{minipage}[b]{1.75cm}
    \scriptsize\centering
    \texttt{"A black Honda motorcycle parked in front of a garage."}
    \end{minipage} & 
    \begin{minipage}[b]{1.75cm}
    \scriptsize\centering
    \texttt{"A room with blue walls and a white sink and door."}
    \end{minipage} &
    \begin{minipage}[b]{1.75cm}
    \scriptsize\centering
    \texttt{"A car that seems to be parked illegally behind a legally parked car"}
    \end{minipage} & 
    \begin{minipage}[b]{1.75cm}
    \scriptsize\centering
    \texttt{"A large passenger airplane flying through the air."}
    \end{minipage} & 
    \begin{minipage}[b]{1.75cm}
    \scriptsize\centering
    \texttt{"The home office space seems to be very cluttered."}
    \end{minipage} & 
    \begin{minipage}[b]{1.75cm}
    \scriptsize\centering
    \texttt{"A cute kitten is sitting in a dish on a table."}
    \end{minipage} & 
    \begin{minipage}[b]{1.75cm}
    \scriptsize\centering
    \texttt{"An open food container box with four unknown food items."}
    \end{minipage} \\ [1ex] 
    
    % Second row of images with red text at the end
    
    \includegraphics[width=0.12\textwidth]{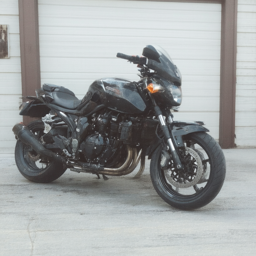} &
    \includegraphics[width=0.12\textwidth]{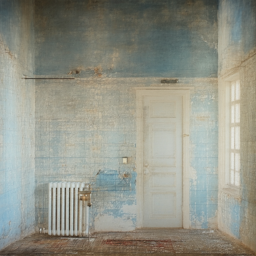} &
    \includegraphics[width=0.12\textwidth]{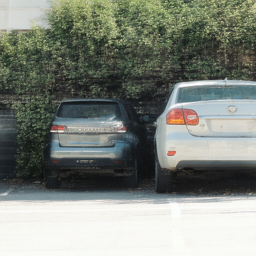} &
    \includegraphics[width=0.12\textwidth]{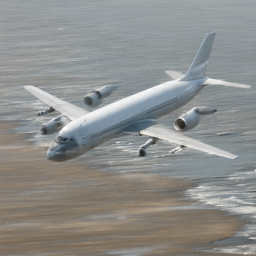} & \includegraphics[width=0.12\textwidth]{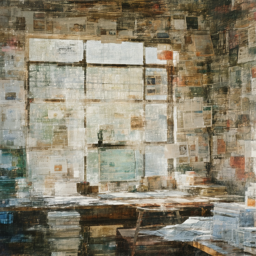}&
    \includegraphics[width=0.12\textwidth]{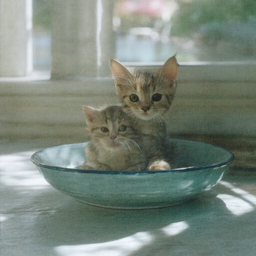}&
    \includegraphics[width=0.12\textwidth]{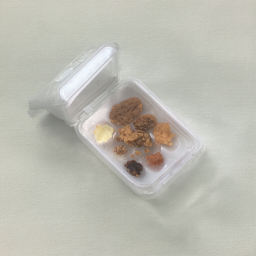}
    \\
    
    \includegraphics[width=0.12\textwidth]{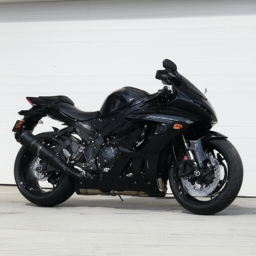} &
    \includegraphics[width=0.12\textwidth]{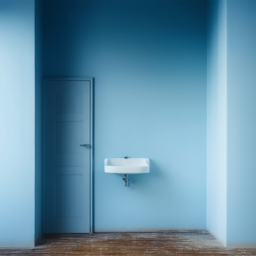} &
    \includegraphics[width=0.12\textwidth]{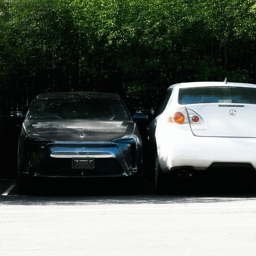} &
    \includegraphics[width=0.12\textwidth]{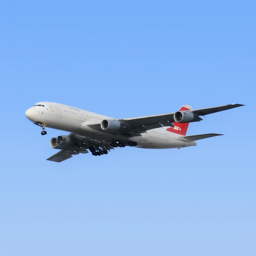} & \includegraphics[width=0.12\textwidth]{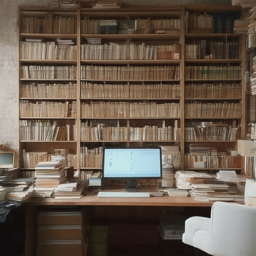}&
    \includegraphics[width=0.12\textwidth]{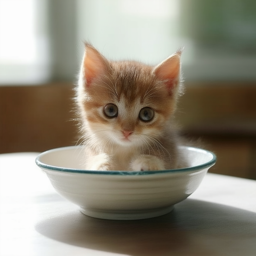}&
    \includegraphics[width=0.12\textwidth]{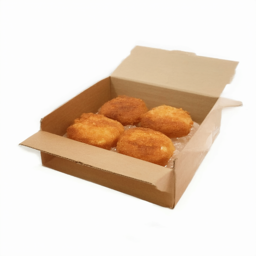}
    \\
  \end{tabular}
  }

  \vspace{3.5ex} % Vertical space between grids

{\setlength{\tabcolsep}{0.5pt}%
  
  \rotatebox{90}{\makebox[0mm][c]{~~\textbf{\underline{FLUX.1-dev}}}}
  \rotatebox{90}{\makebox[0mm][c]{ ~~ Ours ~~~~~~ $\mathcal{N}(0,I)$} ~~~}
  \begin{tabular}{@{}c c c c c c c@{}}
    % Second row of images with red text at the end
    
    \includegraphics[width=0.12\textwidth]{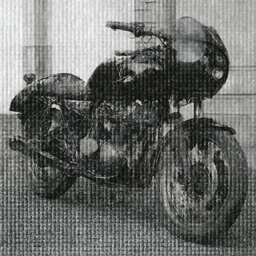} &
    \includegraphics[width=0.12\textwidth]{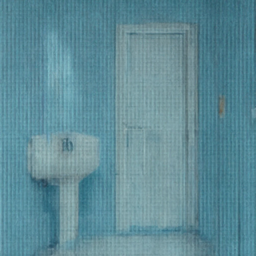} &
    \includegraphics[width=0.12\textwidth]{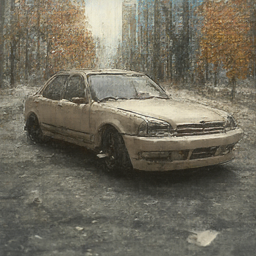} &
    \includegraphics[width=0.12\textwidth]{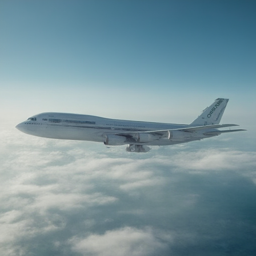}
    & \includegraphics[width=0.12\textwidth]{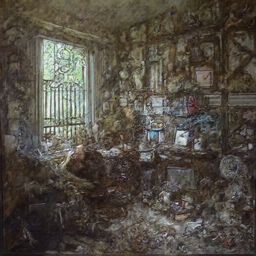}&
    \includegraphics[width=0.12\textwidth]{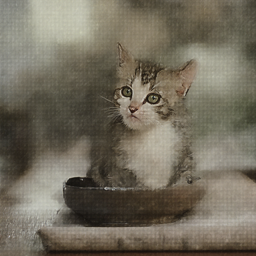}&
    \includegraphics[width=0.12\textwidth]{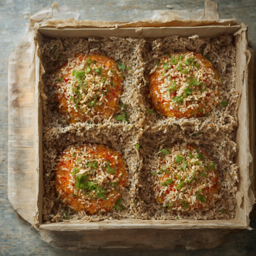}\\
    
    \includegraphics[width=0.12\textwidth]{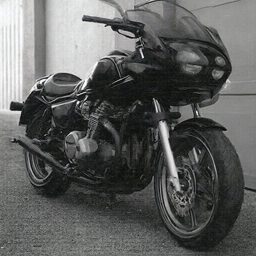} &
    \includegraphics[width=0.12\textwidth]{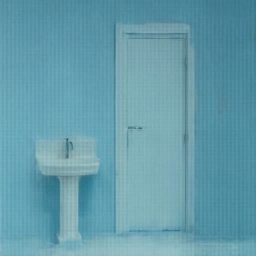} &
    \includegraphics[width=0.12\textwidth]{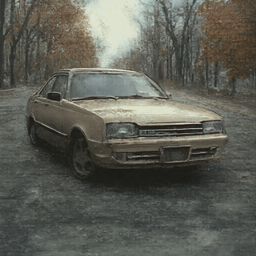} &
    \includegraphics[width=0.12\textwidth]{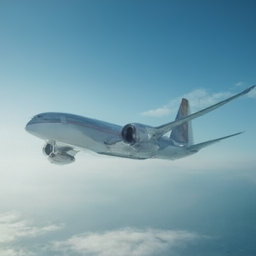}
    &
    \includegraphics[width=0.12\textwidth]{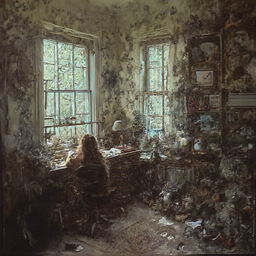}
    &    \includegraphics[width=0.12\textwidth]{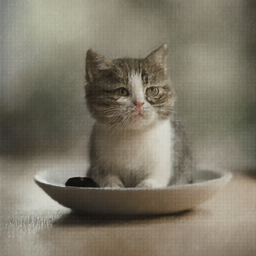}
    &
    \includegraphics[width=0.12\textwidth]{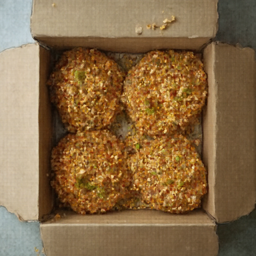}\\
  \end{tabular}
  }
  
   \caption{Qualitative results for Stable Diffusion v3.5 and FLUX.1-dev with and without our NLG.  During generation, no guidance is used ($w=1$).  Prompts used for generation are taken from the MS COCO 2014 validation set and are shown at the top of each column.}
   \label{fig:rectificied_flow_qualitative}
\end{figure*}

\renewcommand{\arraystretch}{1.0}

\subsection{Generalization to Non Text-to-Image Models}
\label{sec:experiments_non_t2i}

We also evaluate our method outside the text-to-image context. At the time of writing, EDM2 \cite{karras2024analyzing} with auto-guidance \cite{karras2024guiding} achieves the state-of-the-art generation performance on ImageNet \cite{deng2009imagenet}.  We reimplement our method using their official codebase, focusing on $512 \times 512$ ImageNet generation with a latent diffusion model \cite{rombach2022high} and UNet \cite{ronneberger2015u} architecture.

We test both unconditional and class conditional generation.  In both cases, we use an EDM2-XS (unconditional) model for $D_0$.  For $D_1$, we use an EDM2-S model, with conditional and unconditional variants being used for their respective tasks. Following EMD2, we measure image quality with both FID and FD\textsubscript{DINOv2} \cite{stein2023exposing}.  For conditional generation, we measure alignment between the generated images and the conditioning class using Top-1 and Top-5 accuracy obtained using a pretrained ResNet-50 \cite{he2016deep}. 50k images are used to calculate evaluation metrics. Like in Section~\ref{sec:experiments_analysis}, we test our approach with guidance (using the preset guidance levels in the official EDM2 codebase) during the generation phase, and without guidance. Given EDM2's larger initial noise variance, we adjust the clipping threshold to $\tau=5.0$ to avoid overly restricting $d$. We use the same $l$ and $s$ parameters as in Section~\ref{sec:experiments_analysis}, with potential for further improvement via hyperparameter tuning.

\begin{table*}
  \begin{center}
  \caption{Results for unconditional and class conditional generation of ImageNet $512\times512$ images, using the state-of-the-art diffusion model (EDM2 \cite{karras2024analyzing}). The column ``AutoG" indicates whether or not autoguidance is applied during the generation phase.}
  \label{tab:edm2}
  \begin{tabular}{|>{\centering\arraybackslash}p{1.1cm}|>{\centering\arraybackslash}p{1.5cm}|>{\centering\arraybackslash}p{1.1cm}|>{\centering\arraybackslash}p{1.5cm}|>{\centering\arraybackslash}p{1.1cm}|>{\centering\arraybackslash}p{1.5cm}|>{\centering\arraybackslash}p{1.5cm}|>{\centering\arraybackslash}p{1.5cm}|}
    \hline 
    \multicolumn{2}{|c|}{} &  \multicolumn{2}{c|}{Unconditional} & \multicolumn{4}{c|}{Conditional}\\
    \hline 
    AutoG & Noise & FID $\downarrow$&FD\textsubscript{DINOv2} $\downarrow$& FID $\downarrow$&FD\textsubscript{DINOv2} $\downarrow$ & \thead{\footnotesize Top-1\\\footnotesize Accuracy $\uparrow$} & \thead{\footnotesize Top-5\\ \footnotesize Accuracy $\uparrow$} \\
    \hline
    \multirow{ 2}{*}{\xmark} & Gaussian & 15.72 & 276.49 & 6.88 & 187.44 & 73.40\% & 90.31\%  \\
     & NLG (Ours) & \textbf{14.50} & \textbf{256.80} & \textbf{5.05} & \textbf{159.47} & \textbf{78.71\%} & \textbf{93.37\%} \\
     \hline
     \multirow{ 2}{*}{\cmark} & Gaussian & \textbf{4.30} & 90.71 & \textbf{2.29} & 88.79 & 87.78\% & 97.47\% \\
     & NLG (Ours) & 4.30 & \textbf{90.20} & 2.38 & \textbf{87.82} & \textbf{88.03\%} & \textbf{97.63\%} \\
    \hline
  \end{tabular}
  \end{center}
  \vspace{-0.3cm}
\end{table*}

Results in Table \ref{tab:edm2} show that, similar to our text-to-image results, our method improves image quality and alignment in the no AutoG guidance setting. When AutoG is used during the generation we also see increases in FD\textsubscript{DINOv2} and classifier accuracy, while these increases are less pleasing.  

These results confirm that our method is effective when no diffusion-level guidance is used during diffusion. This experiment also verifies that our approach is compatible with AutoG, not just CFG. Interestingly, this suggests that the latent space of our weak model $D_0$ can be useful in aligning noise for a different diffusion model $D_1$. Finally, this demonstrates that our method is also capable of aligning noise, even when no condition is given. 

Fig.~\ref{fig:edm2_results} shows images generated with and without guidance for both unconditional and conditional generation on EDM2.  %We see that, in the case where guidance is used during generation, there is not much change in the images.  This is because we use fewer aligning steps when guidance is used in the generation phase.

\subsection{Rectified Flow Models}
\label{sec:rectified_flows}
As shown in \cite{zheng2023guided}, flow models can have a score-function based interpretation, leading to the same guidance formula as diffusion, which our work is based off. We can therefore apply our method to rectified flows without modification. We include our results for rectified flow models Stable Diffusion v3.5 and FLUX.1-dev, which also happen to be SOTA models, in Table \ref{tab:sota_models}. As in Section \ref{sec:experiments_analysis}, we evaluate performance using FID, IS, CLIP Score, ImageReward, HPSv2, and PickScore. For both models we set guidance to $w=1$.  Our method leads to improvements in all metrics across both Stable Diffusion v3.5 and FLUX.1-dev.  This result suggests that our method can be consistently applied across a broad range of models, including both diffusion and rectified flow models. Qualitative results for both Stable Diffusion v3.5 and FLUX.1-dev can be seen in Fig.~\ref{fig:rectificied_flow_qualitative}.

\begin{table*}
\footnotesize
\begin{center}
\caption{Results when our method is applied to SOTA rectified flow models Stable Diffusion v3.5 (SD3.5) and FLUX.1-dev (FLUX.1). We generate 10k samples using prompts from MS COCO, with guidance scale set to 1.0 and use 20 inference steps.}
  \label{tab:sota_models}
    % Use tabular, not table*, because this is not a float
    \begin{tabular}{|p{1.3cm}|>{\centering\arraybackslash}p{1.5cm}|>{\centering\arraybackslash}p{1.2cm}|>{\centering\arraybackslash}p{1.2cm}|>{\centering\arraybackslash}p{1.5cm}|>{\centering\arraybackslash}p{1.9cm}|>{\centering\arraybackslash}p{1.2cm}|>{\centering\arraybackslash}p{1.5cm}|}
    \hline 
    Model & \thead{\footnotesize Aligning \\ \footnotesize Steps} &FID $\downarrow$&IS $\uparrow$& \thead{\footnotesize CLIP \\ \footnotesize Score $\uparrow$} & ImageReward $\uparrow$ & HPSv2 $\uparrow$ & PickScore $\uparrow$  \\
    \hline
    \multirow{ 2}{*}{SD3.5} & 0 &  31.30 & 30.82 & 30.88 & 0.356 & 0.2386 & 21.64 \\
     & 20 & \textbf{25.80} & \textbf{36.18} & \textbf{31.47} & \textbf{0.525} & \textbf{0.2546} & \textbf{22.02} \\
     \hline
     \multirow{ 2}{*}{FLUX.1} & 0 & 51.34 & 24.80 & 29.50 & 0.0475 & 0.2214 & 21.21 \\
     & 10 & \textbf{42.95} & \textbf{27.01} & \textbf{29.86} & \textbf{0.206} & \textbf{0.2272} & \textbf{21.40}   \\
    \hline
  \end{tabular}
  \end{center}
  
\end{table*}

\renewcommand{\arraystretch}{1.0}

\subsection{Applications}\label{sec:nlg_applications}
\begin{figure}[htbp]
   \begin{center}
  \vspace{1mm}
  {\setlength{\tabcolsep}{0.0pt}%
  \begin{tabular}{@{}p{2cm} c c c@{}}
    % Column headers (first empty cell for the row-end text)
     \textbf{Prompt} & \textbf{s=0} & \textbf{s=5} & \textbf{s=10} \\[1ex]
    % First row of images with green text at the end
    \scriptsize \texttt{"There is a two level tour bus in the street."}
    &
    \includegraphics[width=0.11\textwidth,valign=c]{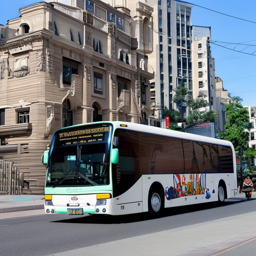} & 
    \includegraphics[width=0.11\textwidth,valign=c]{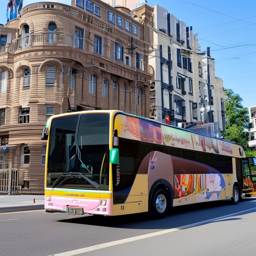} & 
    \includegraphics[width=0.11\textwidth,valign=c]{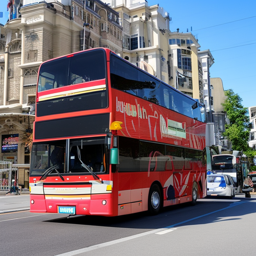} 
    \\
    \scriptsize
    \texttt{"A Miami Air airplane flies against a blue sky."}
    &
    \includegraphics[width=0.11\textwidth,valign=c]{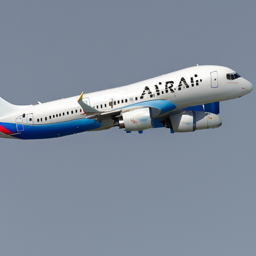} & 
    \includegraphics[width=0.11\textwidth,valign=c]{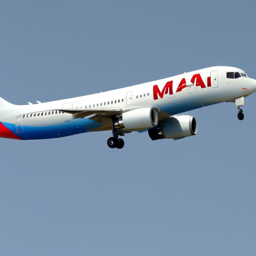} &
    \includegraphics[width=0.11\textwidth,valign=c]{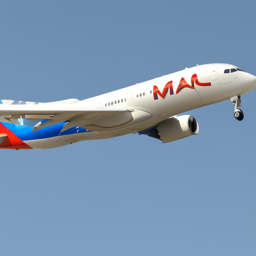}  \\
    \scriptsize
    \texttt{"very well cooked food with raw onions on it"}
    &
    \includegraphics[width=0.11\textwidth,valign=c]{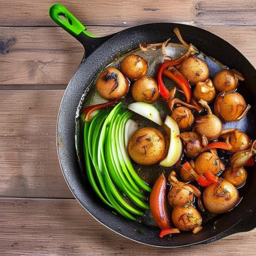} & 
    \includegraphics[width=0.11\textwidth,valign=c]{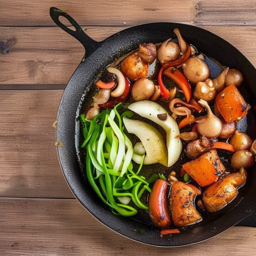} &
    \includegraphics[width=0.11\textwidth,valign=c]{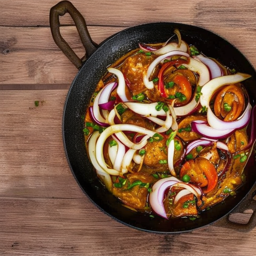}  \\
  \end{tabular}
  }
  \end{center}
   \caption{When guidance is used ($w=7.5$), our NLG method can make images which are poorly aligned with their input prompt, better aligned. $s$ corresponds to the number of aligning steps.  Images generated with Stable Diffusion v2.1.}
   \label{fig:badly_aligned_images}
\end{figure}

\noindent\textbf{{Poorly Aligned Images.}}
InitNO found that, even for high guidance scales, text-to-image models sometimes fail to generate images aligned with their input prompts.  They attribute this problem to the initial noise.  We therefore select poorly aligned images (low CLIP Score) generated with CFG ($w=7.5$) and show the effect of NLG on these images in Fig.~\ref{fig:badly_aligned_images}.

\renewcommand{\arraystretch}{0.25}
\renewcommand{\arraystretch}{1}

\begin{figure}[t]
   \begin{center}
    \includegraphics[width=0.6\linewidth]{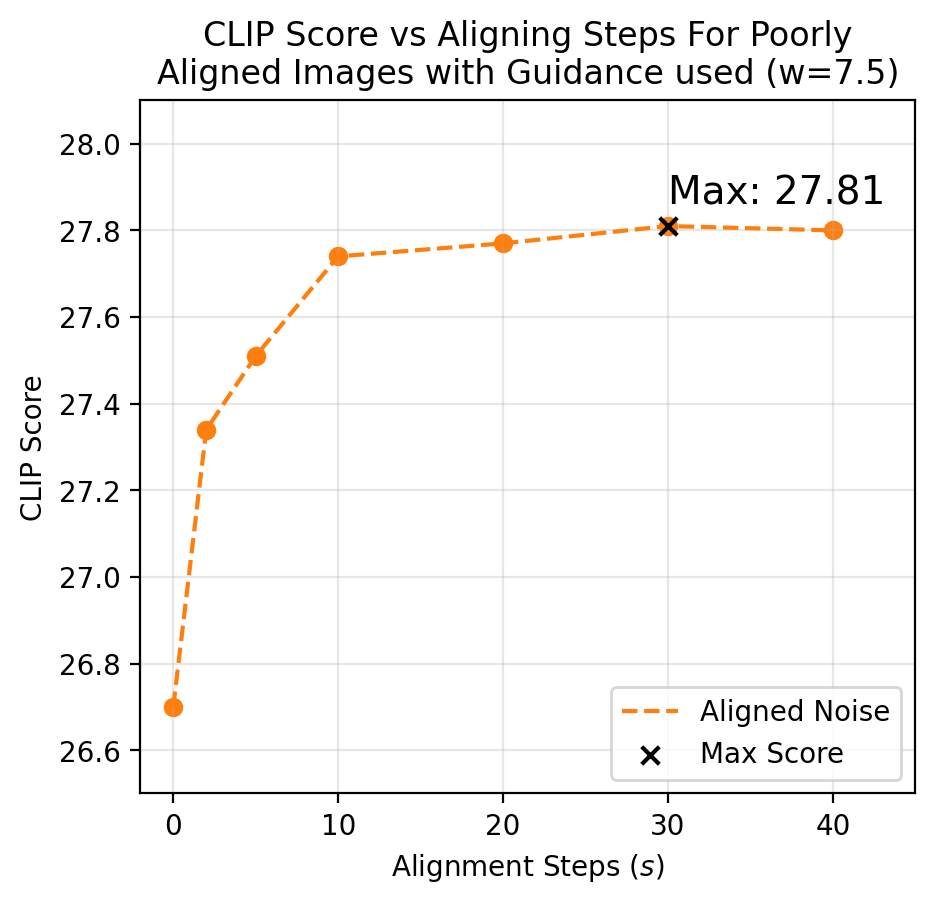}
    \end{center}
\caption{This plot records the CLIP Scores for the 1k worst prompt-aligned (out of 10k) images generated on Stable Diffusion v2.1 with guidance set 7.5.  Prompt-alignment is measured with CLIP Score.  For this subset of images, applying NLG for more and more steps increases the CLIP Score.}
   \label{fig:badly_aligned_plot}
\end{figure}

We can see from this figure that for images with poor prompt alignment, NLG can be an effective tool for improving the image-prompt correspondence.  Therefore, it seems that for a subset of images, noise alignment can be helpful.  To demonstrate this, we generate 10k images with Gaussian noise and a guidance scale of $w=7.5$, using prompts from MS COCO. We then filter for the 1000 images with the lowest CLIP Scores.  In Fig.~\ref{fig:badly_aligned_plot} we compare these low scoring images both with and without NLG. While noise alignment is generally not helpful when using CFG (as seen in Section \ref{sec:experiments_analysis}), Fig.~\ref{fig:badly_aligned_plot} shows that for poorly aligned images, NLG can significantly improve CLIP Score.

\begin{figure}[t]
   \begin{center}

    \texttt{PROMPT: "a portrait photo of an} \\ \texttt{sleepy facial expression"} \\
    \vspace{2mm}
    \includegraphics[width=0.8\linewidth]{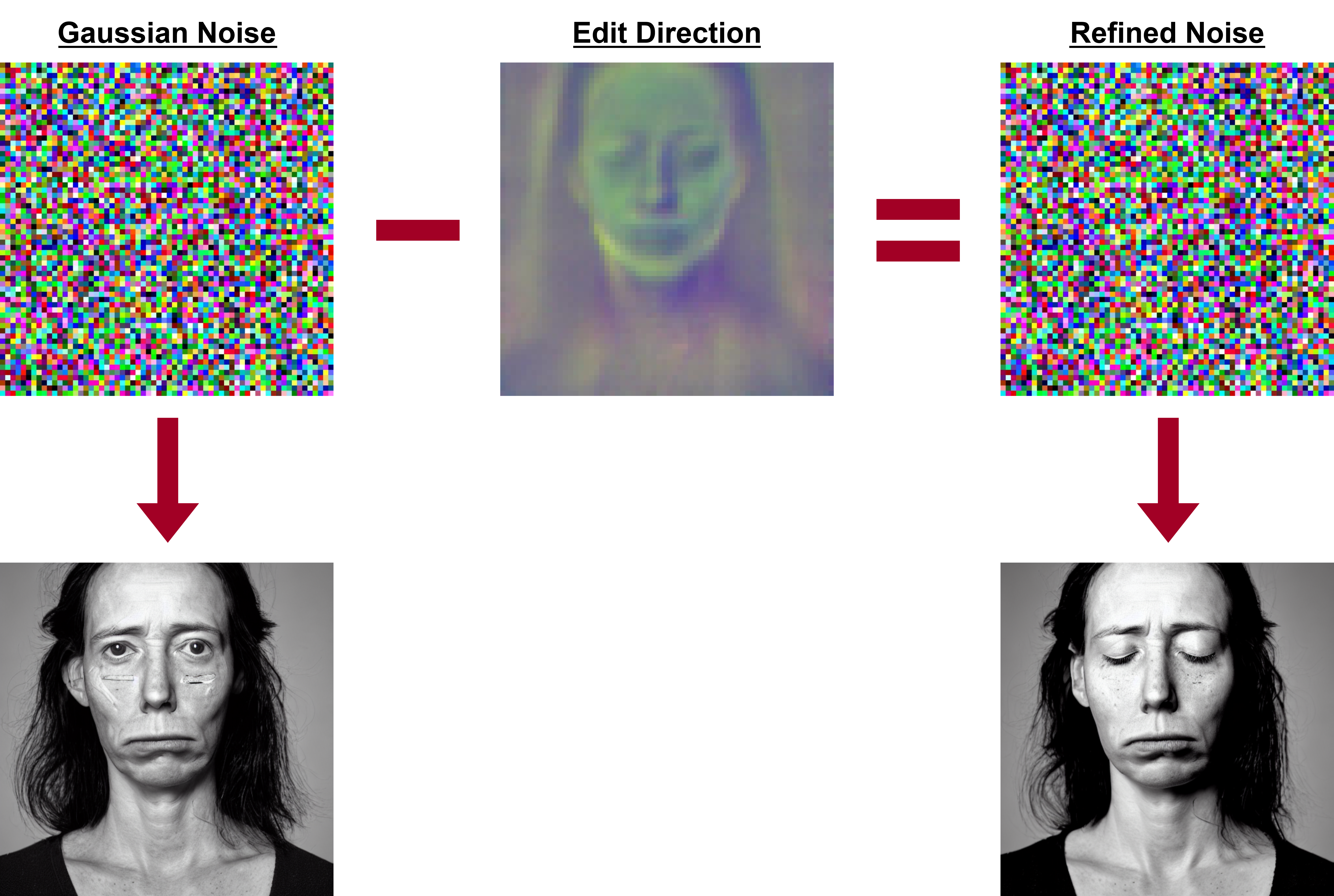}
    \end{center}
\caption{{High level overview of our NLG method.  By setting the noise level to to be $l=0$, we can clearly see the semantics of the edit direction. Images generated using Stable Diffusion v1.5.}}
   \label{fig:method_core_alt}
\end{figure}

\begin{figure}[htbp]
   \centering
  \vspace{1mm}
  {\setlength{\tabcolsep}{4.0pt}%
  \begin{tabular}{@{}p{1.5cm} c : c c@{}}
    % Column headers (first empty cell for the row-end text)
    \textbf{Generation Prompt} & & \multicolumn{2}{c}{\textbf{Aligning Prompt}} \\
      &  & \begin{minipage}[b]{1.5cm} \scriptsize\centering \texttt{"a grid of images"} \end{minipage} & \begin{minipage}[b]{1.5cm} \scriptsize\centering \texttt{"a full moon"} \end{minipage} \\[1ex]
    % First row of images with green text at the end
     \begin{minipage}[b]{1.5cm}\scriptsize \texttt{"a painting of kings and queens"} \end{minipage}
    &
    \includegraphics[width=0.10\textwidth]{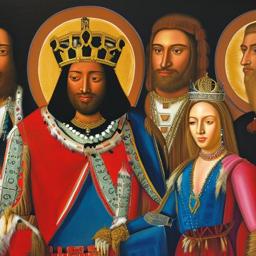} & 
    \includegraphics[width=0.10\textwidth]{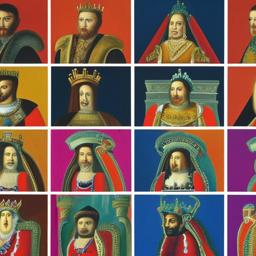} & 
    \includegraphics[width=0.10\textwidth]{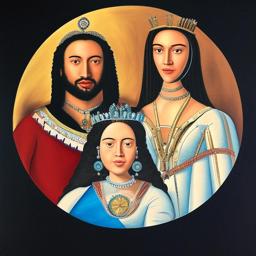} 
    \\
    \begin{minipage}[b]{1.5cm}\scriptsize \texttt{"cats and dogs"} \end{minipage}
    &
    \includegraphics[width=0.10\textwidth]{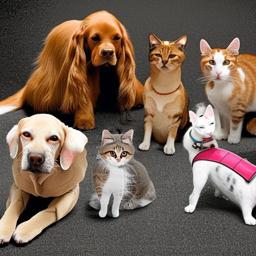} & 
    \includegraphics[width=0.10\textwidth]{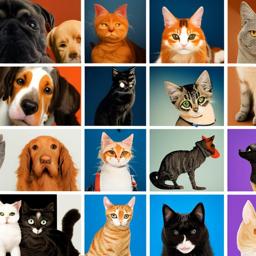} &
    \includegraphics[width=0.10\textwidth]{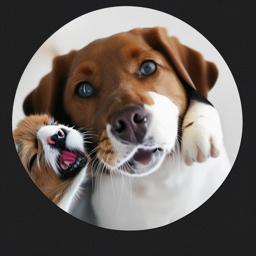}  \\
  \end{tabular}
  }
   \caption{Images are generated using Stable Diffusion v2.1. Generation Prompt is used for CFG, and Aligning Prompt is utilized for NLG. The left column is generated from normal noise.  The middle and right columns use aligned noise.  All generations use the same seed.  The prompt used for NLG is different, for example the bottom right image uses the prompt ``a full moon'' for alignment and the prompt ``cats and dogs'' for generation. }
\label{fig:structuring_noise}
\end{figure}

\noindent\textbf{{Structuring Initial Noise.}} In Fig.~\ref{fig:method_core_alt} we show an overview of our method with $l=0$ for clear visualization.  This allows us to see that our approach adds structure to noise which can be useful for subsequent generation.  We can exploit this by using one prompt during noise alignment to structure the noise and another during generation to fill in the image semantics.   For example, the prompt ``a grid of images'' can be used to structure the noise, followed by ``cats and dogs'' to fill in the content, as shown in Fig.~\ref{fig:structuring_noise}. 

\renewcommand{\arraystretch}{2}

\begin{figure}[htbp]
   \centering
  {\setlength{\tabcolsep}{0.0pt}%
  \begin{tabular}{@{}p{1.1cm} c c c c@{}}
    % Column headers (first empty cell for the row-end text)
     & \scriptsize \begin{minipage}[c]{1.1cm}{Gaussian Noise}\end{minipage} &  \scriptsize \begin{minipage}[c]{1.1cm}{Aligned w/ SD1.5}\end{minipage} & \scriptsize \begin{minipage}[c]{1.1cm}{Aligned w/ SD-Turbo}\end{minipage} & \scriptsize \begin{minipage}[c]{1.1cm}{Aligned w/ EDM2}\end{minipage} \\[1ex]
    % First row of images with green text at the end
    \raggedright\scriptsize Gen w/ SD1.5
    &
    \includegraphics[width=0.10\textwidth,valign=c]{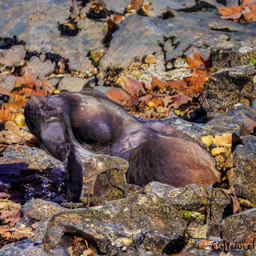} & 
    \includegraphics[width=0.10\textwidth,valign=c]{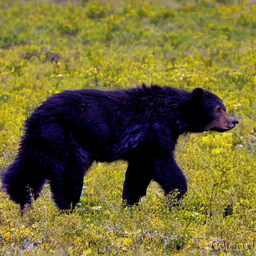} & 
    \includegraphics[width=0.10\textwidth,valign=c]{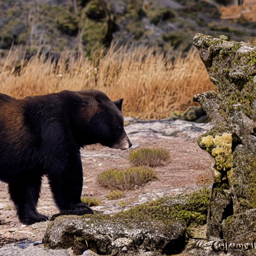} & 
    \includegraphics[width=0.10\textwidth,valign=c]{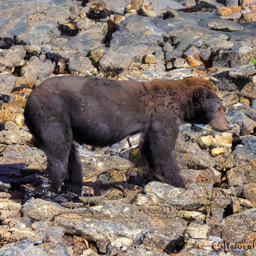} 
    \\
    \raggedright\scriptsize Gen w/ SD-Turbo
    &
    \includegraphics[width=0.10\textwidth,valign=c]{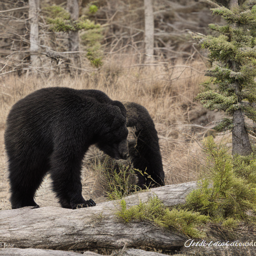} & 
    \includegraphics[width=0.10\textwidth,valign=c]{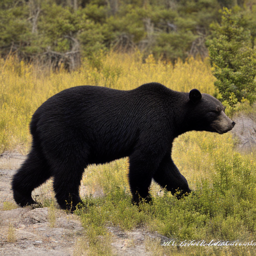} & 
    \includegraphics[width=0.10\textwidth,valign=c]{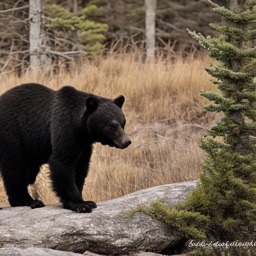} & 
    \includegraphics[width=0.10\textwidth,valign=c]{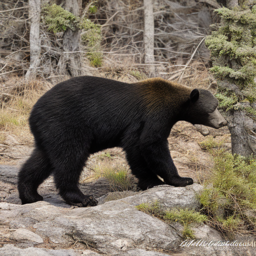} 
    \\
    \raggedright\scriptsize Gen w/ EDM2
    &
    \includegraphics[width=0.10\textwidth,valign=c]{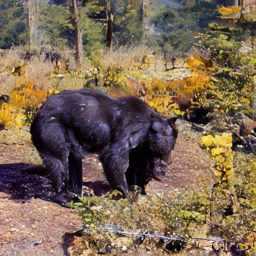} & 
    \includegraphics[width=0.10\textwidth,valign=c]{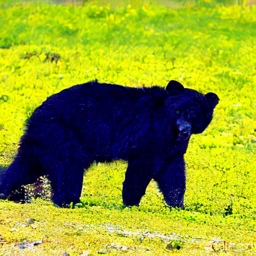} & 
    \includegraphics[width=0.10\textwidth,valign=c]{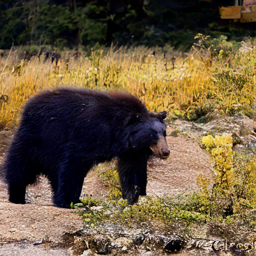} & 
    \includegraphics[width=0.10\textwidth,valign=c]{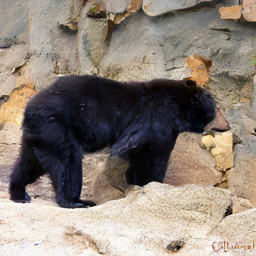} 
    \\
  \end{tabular}
  }
  
   \caption{Images generated across models with the same noise show similar structure.  Images in each column uses the same starting noise which has been aligned using a different diffusion model. Each row uses a different model for generation.  The models used are Stable Diffusion v1.5, SD-Turbo, or EDM2. The text prompt used was ``American black bear, black bear, Ursus americanus, Euarctos americanus'', corresponding to the ImageNet class of the same name. $w=1$ in CFG/AutoG for all cases.}
   \label{fig:cross_model_noise}
\end{figure}

This can also allow aligned noise to transfer between different diffusion models. We can align noise using one diffusion model, and then run inference using another. We have seen how NLG can add structured noise, resulting in similar structures in generated images. Fig.~\ref{fig:cross_model_noise} shows that noise aligned using SD-Turbo \cite{sauer2024adversarial} improves generation on both Stable Diffusion v1.5 and vice versa.  This suggests that NLG, can be applied across and between different diffusion models.

\renewcommand{\arraystretch}{1}

\section{Conclusion}
\label{sec:conclusion}

We introduce noise-level guidance, a simple and computationally efficient method for improving the initial noise in diffusion models. By leveraging linear combinations of the model’s own outputs, our approach identifies an edit direction that, when iteratively applied, can enhance both image quality and alignment with conditioning signals. Unlike prior NLO methods, our technique requires no dataset construction, auxiliary networks, additional training, candidate sampling, or backpropagation, making it significantly more lightweight and efficient. Additionally, it generalizes well to different types of diffusion guidances like CFG and AutoG, as well as both conditional and unconditional generation. While we have demonstrated that our guidance can still be beneficial at high diffusion-level guidance and bring new functions like structuring initial noise and crossing diffusion models, future work will focus on exploring additional advantages of our approach to initial noise optimization.

\section*{Acknowledgments}

This research was mainly supported by Studentship from the Faculty of Engineering and Physical Sciences - Electronic and Computer Science at the University of Southampton. The authors acknowledge the use of the IRIDIS X High Performance Computing Facility and the Southampton-Wolfson AI Research Machine (SWARM) GPU cluster, funded by the Wolfson Foundation, together with the associated support services at the University of Southampton in the completion of this work.

%{\appendix[Proof of the Zonklar Equations]
%Use $\backslash${\tt{appendix}} if you have a single appendix:
%Do not use $\backslash${\tt{section}} anymore after $\backslash${\tt{appendix}}, only $\backslash${\tt{section*}}.
%If you have multiple appendixes use $\backslash${\tt{appendices}} then use $\backslash${\tt{section}} to start each appendix.
%You must declare a $\backslash${\tt{section}} before using any $\backslash${\tt{subsection}} or using $\backslash${\tt{label}} ($\backslash${\tt{appendices}} by itself
% starts a section numbered zero.)}


\bibliography{UOS}

\bibliographystyle{IEEEtran}

\vfill

\newpage

\appendices 

\section{Additional Method Details}

The supplementary material provides additional discussions on the proposed NLG approach, along with extra experimental results. To further illustrate the simplicity of the proposed approach, we include a pseudocode implementation below. Note that the used UNet in NLG is an intrinsic component of the diffusion model, so we do not consider it an auxiliary network. We merely used the pre-trained UNet for predicting noise residual. We are committed to making our research reproducible and accessible to the broader community. Therefore, we will make the complete codebase publicly available as open-source resources.

\definecolor{codegreen}{rgb}{0,0.6,0}
\definecolor{codegray}{rgb}{0.5,0.5,0.5}
\definecolor{codepurple}{rgb}{0.58,0,0.82}
\definecolor{backcolour}{rgb}{0.95,0.95,0.92}

\lstdefinestyle{mystyle}{
    backgroundcolor=\color{backcolour},   
    commentstyle=\small\color{codegreen},
    keywordstyle=\color{magenta},
    numberstyle=\tiny\color{codegray},
    stringstyle=\color{codepurple},
    basicstyle=\ttfamily\footnotesize,
    breakatwhitespace=false,         
    breaklines=true,                 
    captionpos=b,                    
    keepspaces=true,                 
    %numbers=left,                    
    numbersep=5pt,                  
    showspaces=false,                
    showstringspaces=false,
    showtabs=false,                  
    tabsize=2
}

\lstset{style=mystyle}

\begin{lstlisting}[language=Python]
def align_noise(noise, prompt, unet):
    
    # Get latent dimensions
    b, c, w, h = noise.shape

    for _ in range(20):

        # Predict the noise residual
        pred_cond = unet(noise, prompt, t=999)
        pred_uncond = unet(noise, "", t=999)

        # Find the direction
        d = pred_cond-pred_uncond

        # Prevent OOD errors
        d = clip_magnitude(d, 0.5)
        noise -= d-(randn(b,c,w,h)*sqrt(0.001))
        noise = (noise/norm(noise))*sqrt(c*w*h)

    return noise
\end{lstlisting}

\section{Additional Results}

\subsection{Further Comparisons}

This section extends our Competing Methods section, with additional results and experiments on competing noise alignment methods.

\paragraph{InitNO} 

Since our approach provides advantages primarily in scenarios where no guidance is used during generation, our InitNO comparison used a guidance scale of $w=1.0$. For completeness, we also perform benchmarking using Stable Diffusion's default guidance scale of $w=7.5$.  Besides the guidance scale, we use all the same hyperparameters as before including $s=20$ for our NLG. Results for this are shown in Table~\ref{tab:initno_2}.

In the main body of the paper, we compare to InitNO, reporting metrics on the dataset from Attend-and-Excite \cite{chefer2023attend}.  However, this dataset has three subsets: Animal-Animal, Animal-Object, and Object-Object.  For a more fine-grained comparison, we therefore report Image-Text and Text-Text Similarity metrics for each subset of the dataset in Table~\ref{tab:initno_finegrained} for both the no guidance and guidance settings.  We see that our method particularly excels in the Object based subsets.

\begin{figure}[htbp]
   \begin{center}
  \vspace{1mm}
  {\setlength{\tabcolsep}{0.5pt}%
  \rotatebox{90}{\makebox[0mm][c]{Ours ~~~~~~~~~~ InitNO ~~~~~~~~~ SD1.5 ~~~~~~~~}}
  \begin{tabular}{@{}c c c c@{}}
    % Column headers (first empty cell for the row-end text)
    \begin{minipage}[b]{1.8cm}
    \scriptsize\centering\texttt{"a cat and a dog"}\end{minipage} & \begin{minipage}[b]{1.8cm}
    \scriptsize\centering\texttt{"a dog and a pink bench"}\end{minipage} & \begin{minipage}[b]{1.8cm}
    \scriptsize\centering\texttt{"a blue balloon and a orange bench"}\end{minipage} \\[1ex]
    % First row of images with green text at the end
    \includegraphics[width=0.125\textwidth]{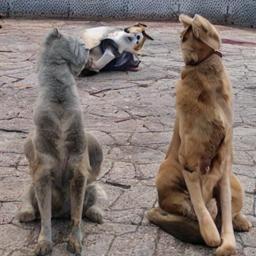} &
    \includegraphics[width=0.125\textwidth]{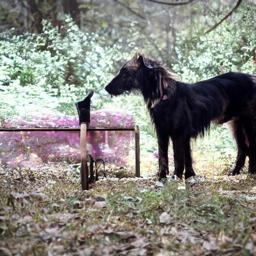}&
    \includegraphics[width=0.125\textwidth]{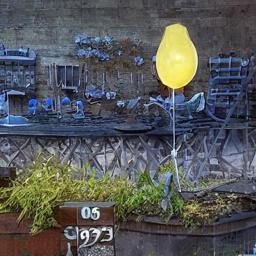}  \\
    \includegraphics[width=0.125\textwidth]{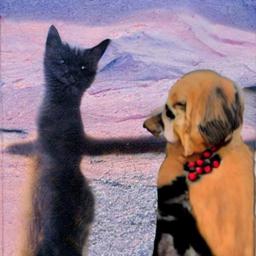} &
    \includegraphics[width=0.125\textwidth]{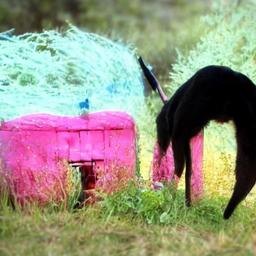}&
    \includegraphics[width=0.125\textwidth]{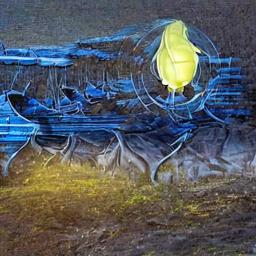} 
    \\
    \includegraphics[width=0.125\textwidth]{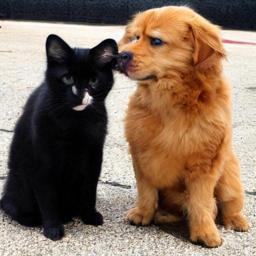} &
    \includegraphics[width=0.125\textwidth]{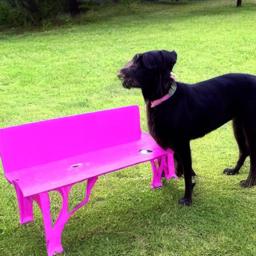}&
    \includegraphics[width=0.125\textwidth]{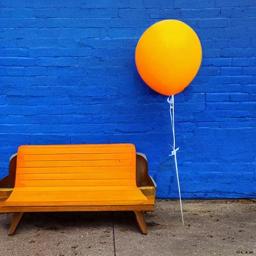} 
  \end{tabular}
  }
   \end{center}
   \caption{Qualitative comparison between our NLG, InitNO, and Stable Diffusion v1.5.  Prompts are taken from the Animal-Animal, Animal-Object, and Object-Object dataset \cite{chefer2023attend}.  Guidance scale for all images is $w=1.0$.}
   \label{fig:qualitative_initno}
\end{figure}

In Fig.~\ref{fig:qualitative_initno} we present a qualitative comparison between our method, InitNO, and Stable Diffusion v1.5, run on the same seeds, using prompts from each subset of the Animal-Animal, Animal-Object, and Object-Object dataset \cite{chefer2023attend}.

\renewcommand{\arraystretch}{0.25}
\renewcommand{\arraystretch}{1}

\begin{table*}
  \begin{center}
  \caption{Comparison with InitNO on the Animal-Animal, Animal-Object, and Object-Object dataset.  This comparison uses CFG while generating images.  Guidance scale is set to 7.5 for all rows. Stable Diffusion v1.5 is used as the base model with 17664 images used to calculate metrics. All experiments are run on an NVIDIA A100 GPU.}
  \label{tab:initno_2}
  \begin{tabular}{|>{\centering\arraybackslash}p{2.0cm}|>{\centering\arraybackslash}p{1.0cm}|>{\centering\arraybackslash}p{1.6cm}|>{\centering\arraybackslash}p{1.6cm}|>{\centering\arraybackslash}p{1.6cm}|>{\centering\arraybackslash}p{1.6cm}|}
    \hline 
Method & IS $\uparrow$ & \shortstack{Text-Text \\ Similarity $\uparrow$} & \shortstack{Image-Text \\ Similarity $\uparrow$} & \shortstack{Generation \\ Time $\downarrow$} & \shortstack{Peak VRAM \\ Usage $\downarrow$} \\    \hline
    Stable Diffusion & 17.33 & 0.77 & 0.33 & 1.7s & 5.3 GB  \\
    InitNO & 14.22 & 0.82 & 0.36 & 12.4s & 17.0 GB   \\
    NLG (Ours) & 17.20 & 0.78 & 0.33 & 3.2s & 5.3 GB \\
    \hline
  \end{tabular}
  \end{center}
\end{table*}

\begin{table*}
  \begin{center}
\caption{CLIP Similarity scores on each subset of the Attend-and-Excite \cite{chefer2023attend} dataset.  We include results for both the no guidance ($w=1.0$) and guidance ($w=7.5$) settings.  Stable Diffusion v1.5 is used as the base model with 17664 images used to calculate metrics.}
  \label{tab:initno_finegrained}
  \begin{tabular}{|>{\centering\arraybackslash}p{1.5cm}|>{\centering\arraybackslash}p{1.5cm}|>{\centering\arraybackslash}p{1.6cm}|>{\centering\arraybackslash}p{1.6cm}|>{\centering\arraybackslash}p{1.6cm}|>{\centering\arraybackslash}p{1.6cm}|>{\centering\arraybackslash}p{1.6cm}|>{\centering\arraybackslash}p{1.6cm}|}
    \hline 
     \multicolumn{2}{|c|}{} &   \multicolumn{2}{c|}{Animal-Animal} & \multicolumn{2}{c|}{Animal-Object} & \multicolumn{2}{c|}{Object-Object} \\
     \hline
Guidance & Method & \shortstack{Text-Text \\ Similarity $\uparrow$} & \shortstack{Image-Text \\ Similarity $\uparrow$} & \shortstack{Text-Text \\ Similarity $\uparrow$} & \shortstack{Image-Text \\ Similarity $\uparrow$} & \shortstack{Text-Text \\ Similarity $\uparrow$} & \shortstack{Image-Text \\ Similarity $\uparrow$} \\
\hline
    
     \multirow{3}{*}{\xmark} & SD1.5 &  0.67 & 0.27 & 0.63 & 0.27 & 0.60 & 0.26 \\
     & InitNO & 0.73 & 0.30 & 0.69 & 0.31 & 0.61 & 0.28 \\
     & NLG (Ours) & 0.71 & 0.30 & 0.74 & 0.32 & 0.74 & 0.32 \\
     \hline
     \multirow{3}{*}{\cmark} & SD1.5 & 0.76& 0.31& 0.78& 0.34& 0.76& 0.34 \\ 
     & InitNO & 0.82& 0.34& 0.83& 0.36& 0.80& 0.36 \\
     & NLG (Ours) & 0.75& 0.31& 0.79& 0.34& 0.78& 0.34 \\
     
    \hline
  \end{tabular}
  \end{center}
\end{table*}

\paragraph{ReNO} Similar to our approach, ReNO repeatedly adjusts noise by finding a direction in latent space.  They find their direction by generating an image, measuring the quality of the image, and then backpropagating through the network.  This restricts ReNO to using 1-step diffusion models because every diffusion step must be backpropagated through.  Our method by comparison is not restricted to 1-step diffusion models, and requires no backpropagation.  Further, measuring the quality of images in ReNO requires at least one additional reward network, whereas our approach requires no additional networks.

\begin{figure}[htbp]
   \begin{center}
  \vspace{1mm}
  {\setlength{\tabcolsep}{0.5pt}%
  \rotatebox{90}{\makebox[0mm][c]{\footnotesize Generated Images ~~~~ Edit Directions ~~~~}}
  \begin{tabular}{@{}c c c@{}}
    % Column headers (first empty cell for the row-end text)
    \footnotesize ReNO & Ours &  \\[1ex]
    % First row of images with green text at the end
    \includegraphics[width=0.13\textwidth]{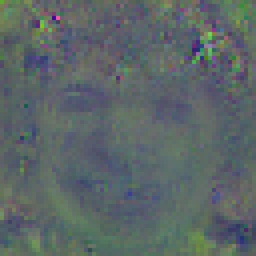} &
    \includegraphics[width=0.13\textwidth]{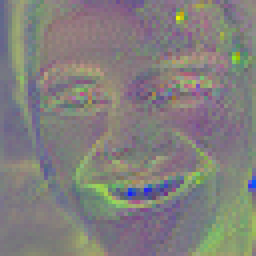}  \\
    \includegraphics[width=0.13\textwidth]{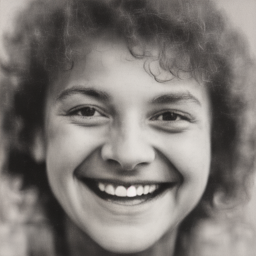} & 
    \includegraphics[width=0.13\textwidth]{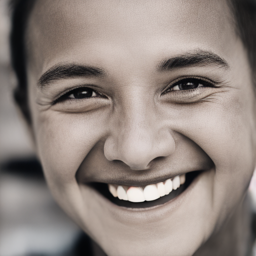} &
    \\
  \end{tabular}
  }
  \end{center}
  \caption{Comparisons of edit directions for ReNO and our NLG method.  Images are generated using SD-Turbo using the same seed. The prompt used was ``a smiling face''.  We also include the generated images.  Both the edit directions and generated images were taken from the 20th step of each algorithm.}
   \label{fig:edit_directions}
   
\end{figure}

In Fig.~\ref{fig:edit_directions} we visualize the edit directions for both ReNO and our approach at the 20th iteration of each algorithm.  We can see that both methods work by introducing structure in to the noise.

We implement our method on SD-Turbo \cite{sauer2024adversarial} to compare against ReNO.  Our method decreases generation time of a single image from 9.23 to 1.11 seconds and decreases peak VRAM usage by 34\% (experiments were run on an NVIDIA A100 GPU).  However, we have found that NLG does not tend to lead improved images for the 1-step diffusion models considered by ReNO.  This is likely because these models are often distilled in such a way that makes CFG much less impactful \cite{meng2023distillation,sauer2024adversarial}. SD-Turbo for example is trained such that during inference CFG not is needed.

\renewcommand{\arraystretch}{0.25}

\begin{figure}[htbp]
   \begin{center}
   \includegraphics[width=0.5\textwidth]{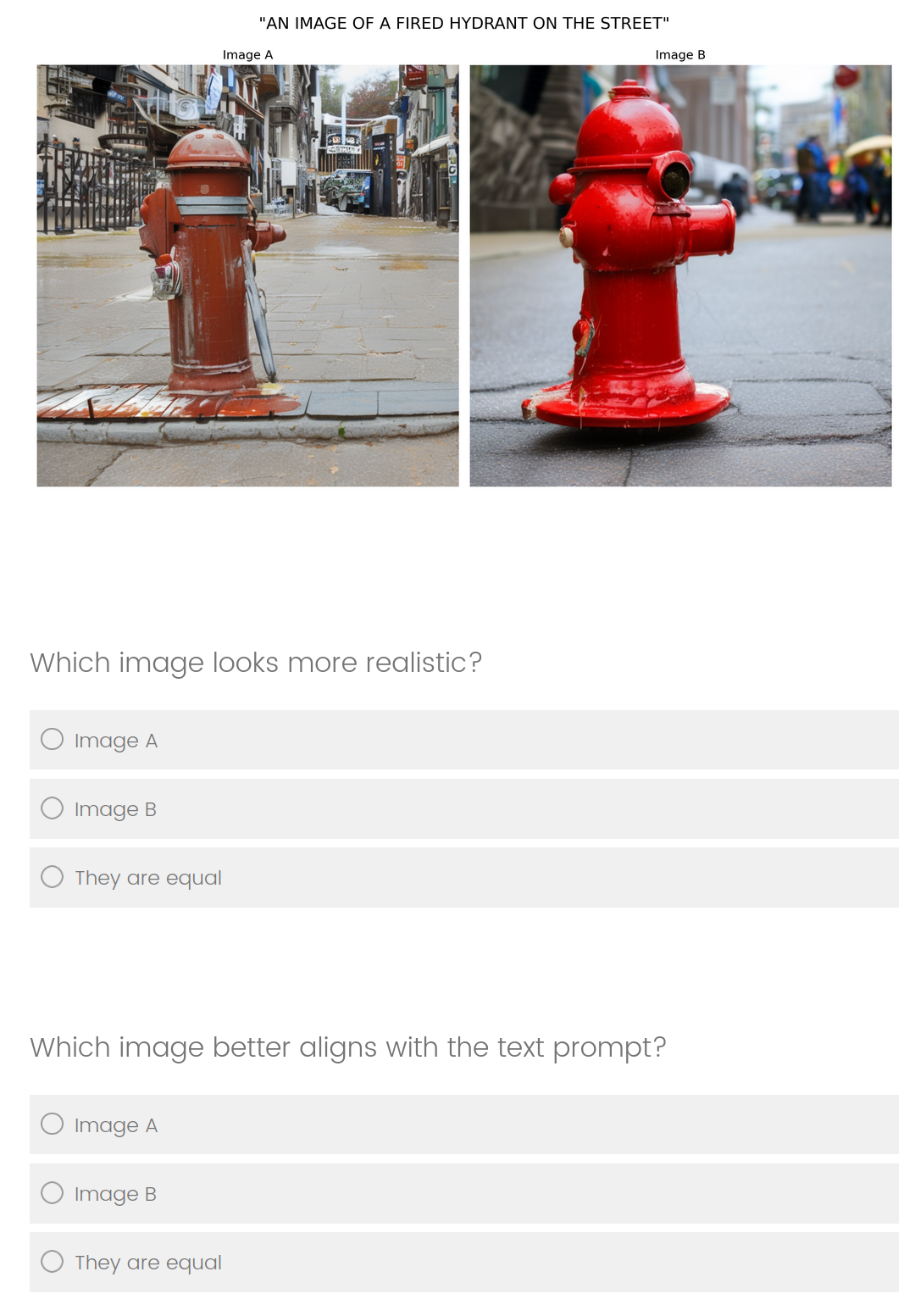}
  \end{center}
  \caption{A screenshot of a single question from our user study taken on qualtrics.com.}
   \label{fig:screenshot}
   
\end{figure}

\subsection{Qualitative Results} 

\begin{figure*}[htbp]
   \centering
  
  %--- Second Grid ---
  \vspace{1mm}
  {\setlength{\tabcolsep}{0.5pt}%

  \rotatebox{90}{\makebox[0mm][c]{ \makebox[0mm][c]{ ~ Ours ~~~~~~~~~ $\mathcal{N}(0,I)$ ~~~~~~~~~~~~~~~~~}}}
  \begin{tabular}{@{}c c c c c c c@{}}
    \begin{minipage}[b]{1.75cm}
    \scriptsize\centering
    \texttt{"A black Honda motorcycle parked in front of a garage."}
    \end{minipage} & 
    \begin{minipage}[b]{1.75cm}
    \scriptsize\centering
    \texttt{"A room with blue walls and a white sink and door."}
    \end{minipage} &
    \begin{minipage}[b]{1.75cm}
    \scriptsize\centering
    \texttt{"A car that seems to be parked illegally behind a legally parked car"}
    \end{minipage} & 
    \begin{minipage}[b]{1.75cm}
    \scriptsize\centering
    \texttt{"A large passenger airplane flying through the air."}
    \end{minipage} & 
    \begin{minipage}[b]{1.75cm}
    \scriptsize\centering
    \texttt{"The home office space seems to be very cluttered."}
    \end{minipage} & 
    \begin{minipage}[b] {1.75cm}
    \scriptsize\centering
    \texttt{"A cute kitten is sitting in a dish on a table."}
    \end{minipage} & 
    \begin{minipage}[b]{1.75cm}
    \scriptsize\centering
    \texttt{"An open food container box with four unknown food items."}
    \end{minipage} \\ [1ex] 
    
    % Second row of images with red text at the end
    
    \includegraphics[width=0.12\textwidth]{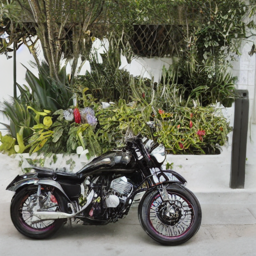} &
    \includegraphics[width=0.12\textwidth]{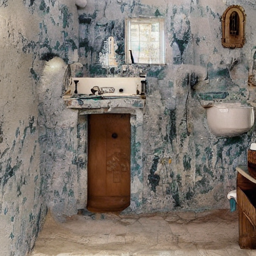} &
    \includegraphics[width=0.12\textwidth]{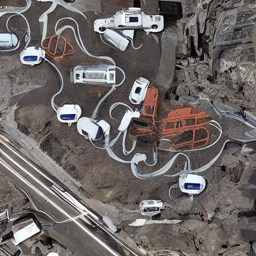} &
    \includegraphics[width=0.12\textwidth]{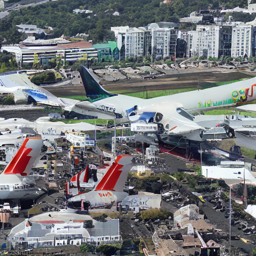} & \includegraphics[width=0.12\textwidth]{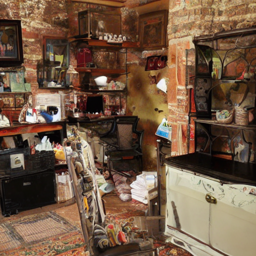}&
    \includegraphics[width=0.12\textwidth]{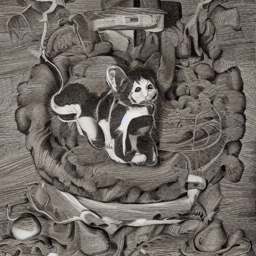}&
    \includegraphics[width=0.12\textwidth]{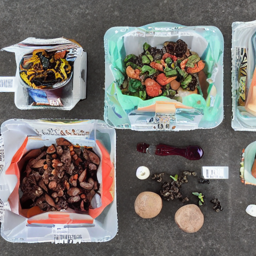}
    \\
    
    \includegraphics[width=0.12\textwidth]{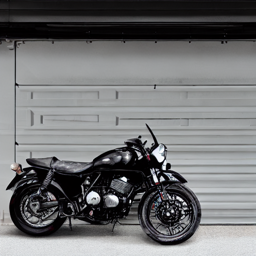} &
    \includegraphics[width=0.12\textwidth]{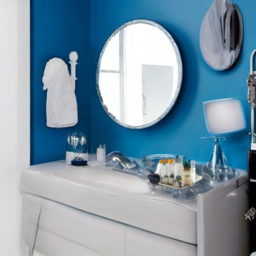} &
    \includegraphics[width=0.12\textwidth]{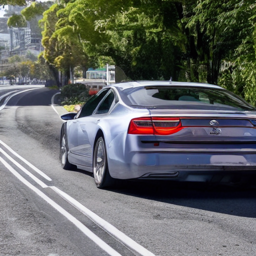} &
    \includegraphics[width=0.12\textwidth]{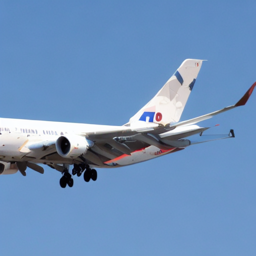} & \includegraphics[width=0.12\textwidth]{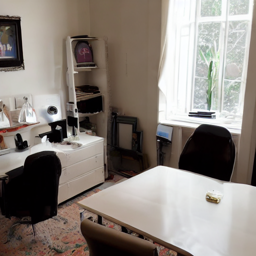}&
    \includegraphics[width=0.12\textwidth]{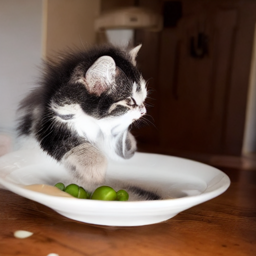}&
    \includegraphics[width=0.12\textwidth]{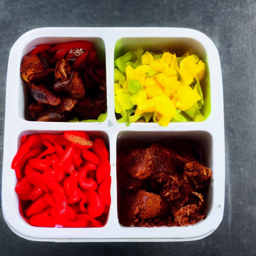}
    \\
  \end{tabular}
  }

   \caption{Qualitative results for Stable Diffusion v1.5 both with and without our NLG.  During generation, no guidance is used ($w=1$).  Prompts used for generation are sourced from MS COCO and are shown at the top of each column.}
   \label{fig:t2i_qualitative_sd1_5}
\end{figure*}

We present qualitative result from our experiments on Stable Diffusion v1.5.  Fig.~\ref{fig:t2i_qualitative_sd1_5} shows images generated with Stable Diffusion v1.5 for MS COCO prompts.  Following our experiments on Stable Diffusion v2.1, we use 20 steps in the no guidance setting.

\renewcommand{\arraystretch}{0.25}

\renewcommand{\arraystretch}{1}

\subsection{Additional User Study Details} 
This section outlines additional details of our user study. Participants were recruited by posting on MS Teams groups at the University. At the start of the survey, participants were given the following instructions: \textit{``...participants will be shown two images and asked to select which one is more realistic. Images may also be generated based on a condition. For example, an AI may use the text prompt ``a photo of an astronaut riding a horse on mars'' to generate a picture. Participants will therefore also be asked which of the two picture better represents this condition''}.  They were then asked to consent to take part in the survey before continuing. We show a screenshot of a question from our survey in Fig.~\ref{fig:screenshot}.

This study was approved by the Faculty Research Ethics Committee (FREC) at the University of Southampton. Data were stored on qualtrics and held in accordance with University policy on data retention.

\end{document}